\pdfoutput=1

\documentclass[11pt]{article}

\usepackage{acl}

\usepackage{times}
\usepackage{latexsym}

\usepackage[T1]{fontenc}

\usepackage[utf8]{inputenc}

\usepackage{microtype}
\usepackage[ruled,linesnumbered]{algorithm2e}
\usepackage{graphicx}
\usepackage{times}
\usepackage{latexsym}
\usepackage{booktabs}
\usepackage{xcolor,colortbl}
\usepackage{diagbox}
\usepackage{graphicx}
\usepackage{amsmath}
\usepackage{xfp}
\usepackage{tabulary}
\usepackage{threeparttablex}
\usepackage{multicol}
\usepackage{array}
\usepackage{amsthm,amsmath,amssymb}
\usepackage{tabularx}
\usepackage{mathrsfs}
\usepackage{multirow}
\usepackage{arydshln}
\usepackage{paralist}
\usepackage{listings}
\usepackage{verbatim}
\usepackage{tcolorbox}

\usepackage{color}

\newcommand{\ourmethod}{\textsc{AutoScraper} }

\newcommand{\orgone}{\textsuperscript{$\diamondsuit$}}
\newcommand{\orgtwo}{\textsuperscript{$\heartsuit$}}
\newcommand{\orgthree}{\textsuperscript{$\spadesuit$}}

\renewcommand{\thefootnote}{\fnsymbol{footnote}}

\title{\textsc{AutoScraper}: A Progressive Understanding Web Agent for Web Scraper Generation}

\author{Wenhao Huang\orgone, Zhouhong Gu\orgone, Chenghao Peng\orgtwo, Zhixu Li\orgone, Jiaqing Liang\orgtwo \footnote[2]{}\\ {\bf Yanghua Xiao\orgone\footnote[2]{}, Liqian Wen\orgthree, Zulong Chen\orgthree}\\
\orgone Shanghai Key Laboratory of Data Science, School of Computer Science, Fudan University \\
\orgtwo School of Data Science, Fudan University,
\orgthree Alibaba Holding-Aicheng Technology-Enterprise \\
\texttt{\{whhuang21,zhgu22,chpeng23\}@m.fudan.edu.cn}, \\
\texttt{\{liangjiaqing,zhixuli,shawyh\}}\texttt{@fudan.edu.cn}
}

\begin{document}
\maketitle
\footnotetext[2]{Corresponding authors.}
\renewcommand*{\thefootnote}{\arabic{footnote}}
\begin{abstract}
Web scraping is a powerful technique that extracts data from websites, enabling automated data collection, enhancing data analysis capabilities, and minimizing manual data entry efforts. Existing methods, wrappers-based methods suffer from limited adaptability and scalability when faced with a new website, while language agents, empowered by large language models (LLMs), exhibit poor reusability in diverse web environments.
In this work, we introduce the paradigm of generating web scrapers with LLMs and propose \textsc{AutoScraper}, a two-stage framework that can handle diverse and changing web environments more efficiently. 
\ourmethod leverages the hierarchical structure of HTML and similarity across different web pages for generating web scrapers. Besides, we propose a new executability metric for better measuring the performance of web scraper generation tasks.
We conduct comprehensive experiments with multiple LLMs and demonstrate the effectiveness of our framework.
Our work is now open-source.\footnote{Resources of this paper can be found at \url{https://github.com/EZ-hwh/AutoScraper}}

\end{abstract}

\section{Introduction}
Web scraping is a process where software automates the extraction of data from websites, typically using bots or web scrapers to gather specific information~\cite{thapelo2021sasscal}. It is important because it allows for efficient data collection and aggregation, which can be crucial for market research, competitive analysis, and real-time data monitoring.

Due to the diversity of sources and information on the internet, the construction of a web scraper requires substantial human effort. Consequently, two types of methods for automatic web information acquisition have been proposed, categorized as wrapper-based and language-agent-based~\cite{sarkhel2023self}.
The wrapper-based method entails complex sequences of operations within customized rule-based functions, which are designed to efficiently access and retrieve desired data from websites, which is especially beneficial for structured websites with stable layouts~\cite{kushmerick1997wrapper,dalvi2011automatic,10.14778/2536206.2536209}. 
Conversely, the language-agent-based method leverages powerful natural language processing capabilities of large language models (LLMs) to interpret free-text queries and directly extract data within websites to meet the demand, effectively handling both structured and dynamic web content~\cite{whitehouse2023webie, scrapegraph-ai}.

\begin{figure}[t]
	\centering
	\includegraphics[width=\columnwidth]{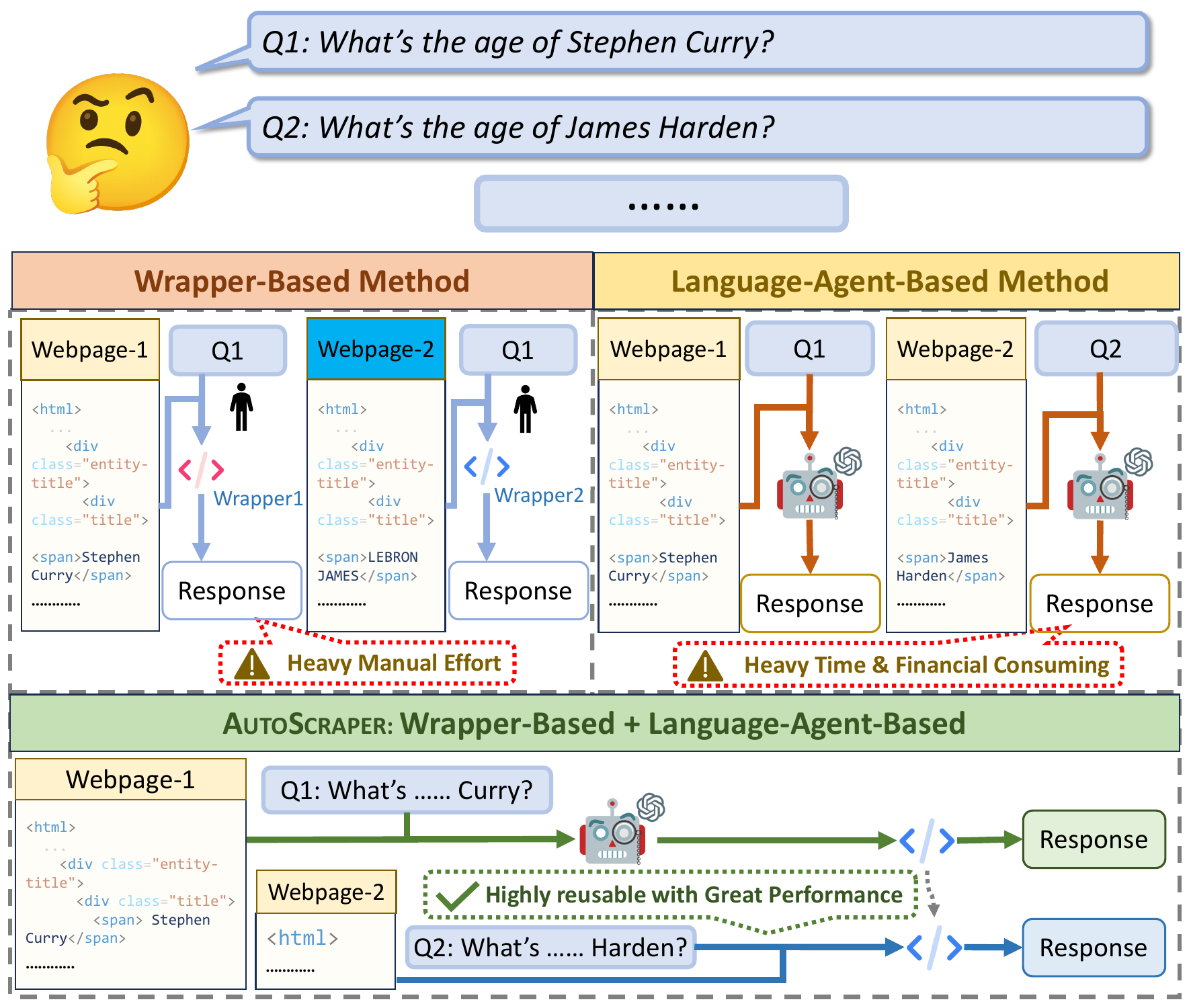}
	\caption{An illustration of comparing wrapper-based methods, language-agent-based methods and \ourmethod.}
	\label{fig:intro}
\end{figure}

Although both types of methods facilitate web scraping to varying degrees, as shown in Figure~\ref{fig:intro}, they exhibit significant shortcomings in terms of scalability.
Wrapper-based method, while reusable, struggles with entirely new website structures, which necessitates extensive human effort to develop additional customized functions~\cite{5767842,lockard2019openceres}.
Conversely, although language-agent-based methods demonstrate superior performance in adapting to new content, their reliance on a limited number of super-powerful API-based LLMs for web scraping incurs considerable time and financial costs.
Together, these challenges impede the broader adoption and scalability of current web scraping technologies, limiting their practicality in dynamic and diverse web environments.

To address the shortcomings of the aforementioned two paradigms, the paradigm of generating web scrapers with LLMs would be the optimal solution. On one hand, compared to wrapper-based methods, it fully leverages the reasoning and reflection capacities of LLMs, reducing manual design on new tasks and enhancing scalability.
On the other hand, compared to language-agent-based methods, it introduces repeatable extraction procedures, reducing the dependency on LLMs when dealing with similar tasks, and thereby improving efficiency when handling a large number of web tasks.
However, there are several challenges associated with using LLMs to generate web scrapers: 
\begin{enumerate}
    \item \textbf{Long HTML document.} Although LLMs excel in comprehending long textual content, HTML, as semi-structured data, comprises both structured (tags and attributes) and unstructured (textual content) elements. Consequently, it is challenging for LLMs to generate executable web scrapers that strictly adhere to the hierarchical structure of web pages in complex markup contexts.
    \item \textbf{Reusability.} A good scraper needs to be reusable across multiple web pages. However, the differences in content and structure between various web pages can lead to the creation of a scraper that references a webpage, which can only be applied to some web pages.
    \item \textbf{Appropriate evaluation metrics.} For a scraper to be considered useful, it must be able to automatically extract the desired results from all web pages. However, existing evaluation metrics for web information extraction, which focus on the extraction results from individual web pages, do not adequately reflect the usability of the scraper. This can potentially mislead experimental conclusions.
\end{enumerate}
We introduce \textsc{AutoScraper}, a two-stage framework to address the web scraper generation task. 
Illustrated in Figure~\ref{fig:framework}, \ourmethod comprises two main components: progressive generation and synthesis.
The progressive generation stage leverages the hierarchical structure of HTML for progressive understanding to address the long HTML document.
Subsequently, the synthesis module integrates multiple scrapers generated on different web pages to produce a cohesive, website-specific scraper that functions universally within that site.
Besides, we propose a new evaluation metric for web scraper generation tasks, called the executability metric. Unlike traditional information extraction metrics that measure single web pages, this metric measures multiple web pages within a website, accurately reflecting the reliability and reusability of the scraper.

We evaluate \ourmethod on three available datasets with 8 LLMs.
On all three datasets, \ourmethod consistently outperforms all baselines and achieves new state-of-the-art results in zero-shot settings.
Also, \ourmethod can surpass supervised learning methods.
Moreover, \ourmethod demonstrates superior efficiency on large-scale web information extraction tasks.
Compared to traditional wrappers, \ourmethod adjusted more quickly according to different websites and task requirements. This flexibility enables scrappers to handle diverse and changing web environments more efficiently. 
Compared to the language agent paradigm, it introduces intermediate functions to enhance reusability and reduce the dependency on LLMs when dealing with similar tasks, thereby improving efficiency when handling a large number of web tasks.

\section{Related Work}
Wrapper-based methods for web scraping utilize the hierarchical structure of the webpage. Method of this category includes rule-based~\cite{zheng2008url}, learning wrappers (i.e a DOM-specific parser that can extract content)~\cite{5767842,kushmerick1997wrapper,dalvi2011automatic}, heuristic algorithm~\cite{lockard2018ceres,lockard2019openceres} and deep learning neural network~\cite{lin2020freedom,zhou2021simplified,li2022markuplm,wang2022webformer}.
These methods demand substantial human involvement, including creating wrapper annotations, applying heuristic scoring rules (such as visual proximity), crafting features for neural network input, and using prior knowledge for verification. Therefore, it is difficult for wrapper-based methods to automatically scale up when facing web scraping tasks across a large number of different websites.

With the emergence of powerful LLMs~\cite{openai2023gpt4,touvron2023llama}, language agents~\cite{sumers2023cognitive} now operate in interactive environments, leveraging LLM-based reasoning, grounding, learning, and decision-making. General language agents, such as Chain-of-Thought~\cite{wei2023chainofthought}, Reflexion~\cite{shinn2023reflexion}, Self-Refine~\cite{madaan2023selfrefineiterativerefinementselffeedback}, and Self-Debug~\cite{chen2023teachinglargelanguagemodels}, capitalize on LLMs' self-reflection capabilities for iterative planning optimization. However, these agents do not effectively utilize web structural features and fail to simplify the web environment after unsuccessful planning attempts, limiting the optimization of subsequent planning.

Current language agents primarily aim to streamline the web environment~\cite{sridhar2023hierarchical, gur2023realworld, zheng2024synapse} and develop strategies for planning and interacting with the web~\cite{sodhi2023heap, ma2023laser}. Nevertheless, these frameworks mainly focus on the concept of open-world web simulation environments~\cite{Shi2017WorldOB, yao2023webshop, deng2023mind2web, zhou2023webarena}, which encompass a broad spectrum of tasks found in real-life scenarios, such as online shopping, flight booking, and software development. These task scenarios are oriented towards individuals and have significantly different requirements for accuracy and efficiency compared to web scraping.

As a result, current language-agent-based methods cannot effectively exploit the HTML structural similarities across multiple web pages, reducing their dependency on LLMs when performing repetitive operations and leading to inefficiencies.

\section{Preliminaries}
\begin{figure*}[t]
    \centering
	\includegraphics[width=\linewidth]{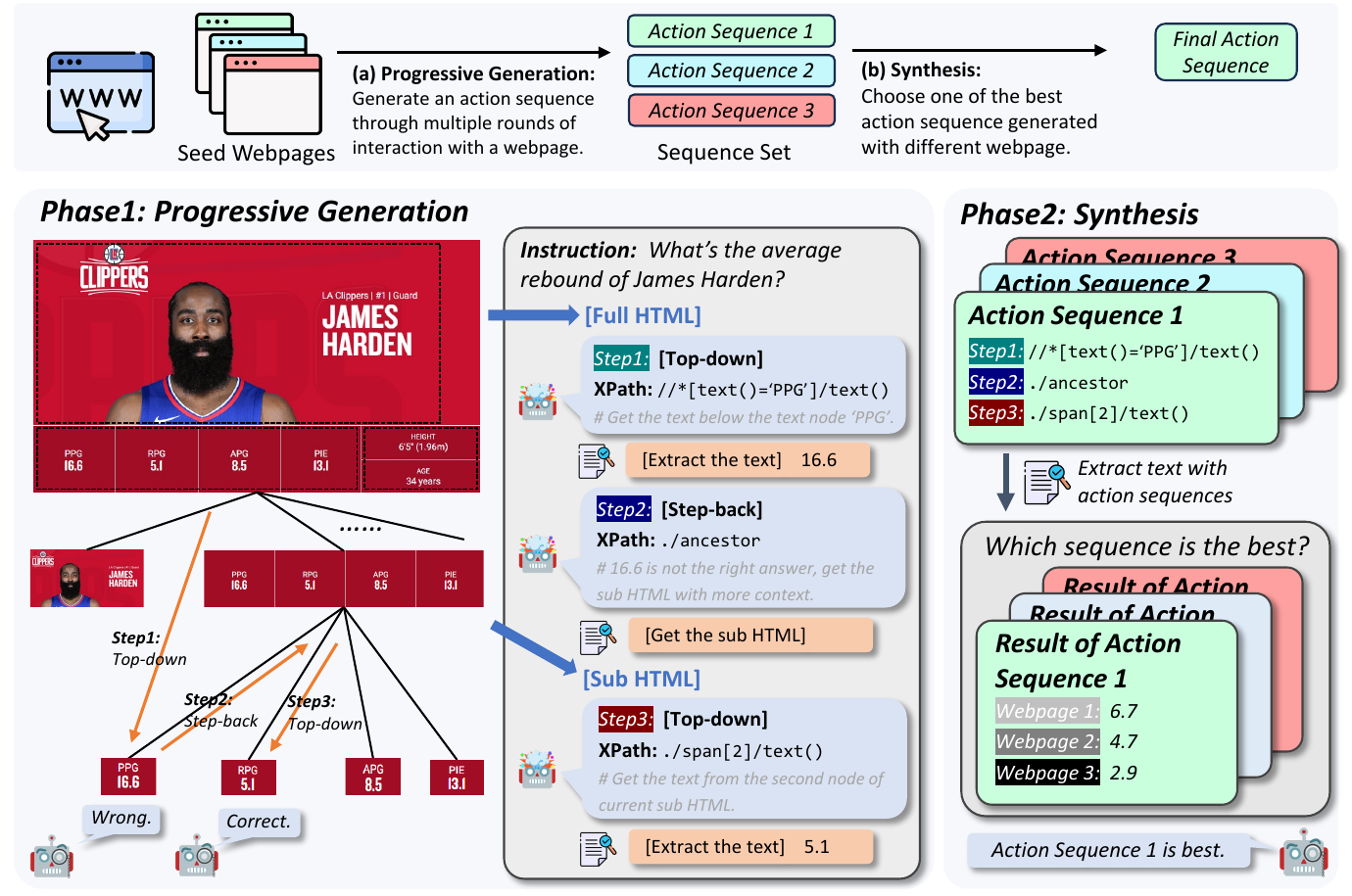}
	\caption{\ourmethod framework of two phases: \textbf{(a) progressive generation} and \textbf{(b) synthesis}.}
	\label{fig:framework}
\end{figure*}

In this section, we first define the scraper generation task and then present the dataset collection process and its corresponding evaluation metrics.

\subsection{Task Formulation}

First, we formulate our scraper generation task. Given a set of webpages on the same website $w\in\mathcal{W}$ describing a subject entity $s$ (also called topic entity in the previous literature), and its corresponding predefined target attribute $r\in\mathcal{R}$, the task objective is to generate an executable rule/action sequence $\mathcal{A}$ to extract target information $o$ from all webpages.

\subsection{Datasets}
We adopt the semi-structure information extraction task as a testbed for the scraper generation task. 

\paragraph{\textsc{Swde}}~\cite{swde_dataset} is a Structured Web Data Extraction dataset that contains webpages from 80 websites in 8 domains, with 124,291 webpages. Each of the websites from the same domains focuses on 3-5 attributes in the web pages. 

\paragraph{\textsc{Extended Swde}}~\cite{lockard2019openceres} involves fine-grained manual annotation of 21 sites in 3 domains from \textsc{Swde}. While \textsc{Swde} contains an average of 4,480 triples for 3 predicates per website, the \textsc{Extended Swde} dataset averages 41K triples for 36 predicates per site.

\paragraph{\textsc{Ds1}}~\cite{omari2017synthesis} contains 166 annotated webpages from 30 real-life large-scale websites categorized into books, shopping, hotels, and movies.

\begin{table}
    \centering
    \small
    \begin{tabular}{lccc}
    \toprule
        \textbf{Dataset} & \textbf{$\text{Num}_\text{Case}$} & \textbf{$\text{Num}_\text{Task}$} & \textbf{$\text{Num}_\text{Web}$}\\
    \midrule
        \textsc{Swde} & 320 & 32 & 32,000\\
        \textsc{Extended Swde} & 294 & 221 &29,400\\
        \textsc{Ds1} & 83 & 11 & 186\\
    \bottomrule
    \end{tabular}
    \caption{The statistic of web scraping task benchmarks. We report the number of the case (\textbf{$\text{Num}_\text{Case}$}), the number of the different extraction task (\textbf{$\text{Num}_\text{Task}$}) and the total number of webpages (\textbf{$\text{Num}_\text{Web}$}).}
    \label{tab:dataset_stat}
\end{table}
We transform the dataset with the following settings.
First, we design instructions for each of the domains, and for each of the attributes as the input information for LLMs\footnote{Further details about the prompt is in Appendix~\ref{apd:task_prompt}}.
Second, for each website in each domain, we sample 100 web pages as the whole test set. 
We consider the set of webpages on the same websites and the corresponding extraction instruction as a case.
For example, for the ESPN websites\footnote{\url{https://global.espn.com/nba/}} in NBA player domains, the sampled 100-detail webpage of players and the instruction \textit{Please extract the team of the player he plays now} is a complete case of our scraper generation task. 
Third, we pre-process the web pages by removing irrelevant elements in a webpage. We use open-source BeautifulSoup library\footnote{\url{https://beautifulsoup.readthedocs.io}} and filter out all DOM element nodes with \texttt{<script>} and \texttt{<style>}, as well as delete all attributes in the element node except \texttt{@class}. We replace the original escape characters in the annotations to ensure consistency with the corresponding information on the web. The statistic of the dataset we transformed is shown in Table~\ref{tab:dataset_stat}.

\subsection{Evaluation Metrics}
\label{sec:evaluation}
Existing evaluation schemes for web page information extraction tasks still follow the traditional metrics of text information extraction tasks, namely precision, recall, and F1 score.
They limit the assessment of methods for the scraper generation task to two aspects. 
First, it focuses on extraction with a single webpage, rather than considering the generalizability from the perspective of a collection of webpages.
Second, it does not effectively measure the transferability when adopting the action sequence to other web pages.

To address this issue, we transform the traditional IE task evaluation into an executable evaluation. 
Based on the traditional IE evaluation on a collection of web pages, we categorize the executability of action sequences into the following six situations. Specifically, for each extraction task on a website, the result is classified based on the extraction result on precision, recall, and f1-score.
\begin{inparaenum}[(1)]
    \item \textbf{Correct}: both precision, recall and f1-score equal 1, which indicates the action sequence is precisely;
    \item \textbf{Precision(Prec.)}: only precision equals 1, which indicates perfect accuracy in the instances extracted following the action sequence, but misses relevant instances;
    \item \textbf{Recall(Reca.)}: only recall equals 1, which means that it successfully identifies all relevant instances in the webpage but incorrectly identifies some irrelevant instances;
    \item \textbf{Un-executable(Unex.)}: recall equals 0, which indicates that the action sequence fails to identify relevant instances;
    \item \textbf{Over-estimate(Over.)}: precision equals 0, which indicates that the action sequence extracts the instances while ground truth is empty;
    \item \textbf{Else}: the rest of the situation, including partially extracting the information, etc.
\end{inparaenum}

Since the above classifications are mutually exclusive, we use the ratio metric to calculate the proportion of each result in our task.
\begin{equation}
    M_R = \frac{\#\ \text{case of situation}}{\#\ \text{total case}}
\end{equation}
We are more concerned with success rate, so for the \textit{Correct} metric, higher values indicate a better proportion of generated execution paths; whereas for the \textit{Un-executable} metric, lower values are preferable.
\section{\ourmethod}

In this section, we describe our framework \ourmethod for generating a scraper to extract specific information from semi-structured HTML.
Our approach is divided into two phases: first, we adopt a progressive generation module that utilizes the hierarchical structure of web pages; second, we employ a synthesis module based on results from multiple web pages. The overall framework is presented in Figure~\ref{fig:framework}.

\subsection{Modeling}
Unlike the wrapper method that generates an XPath, we model the scraper generation task as an action sequence generation task. In specific, we generate an action sequence $\mathcal{A}_{seq}$ that consists of a sequence of XPath\footnote{\url{https://en.wikipedia.org/wiki/XPath}} expression from a set of seed webpages (i.e., a small portion of webpages in the test case for generating the sequence). 
\begin{equation}
    \mathcal{A}_{seq} = [\text{XPath}_1, \text{XPath}_2, ..., \text{XPath}_n]
\end{equation}
where $n$ denotes the length of the action sequence. We execute the XPath in the sequence using the parser in order. 
In the sequence, all XPath expressions except the last one are used for pruning the web page, and the last one is used for extracting the corresponding element value from the pruned web page.

\newcolumntype{o}{>{\columncolor{red!10}}c}
\newcolumntype{b}{>{\columncolor{blue!10}}c}
\begin{table*}[t]
    \centering
    \fontsize{8pt}{9pt}\selectfont
    \setlength{\tabcolsep}{1mm}
    \resizebox{\textwidth}{!}{
        \begin{tabular}{lcoooooobbbooboob}
        \toprule
            \multicolumn{2}{c}{\textbf{Dataset}} & \multicolumn{9}{c}{\textsc{Swde}} & \multicolumn{3}{c}{\textsc{Extended Swde}} & \multicolumn{3}{c}{\textsc{Ds1}}\\
            
            \cmidrule(r){1-2}\cmidrule(r){3-11}\cmidrule(r){12-14}\cmidrule(r){15-17}
            
            \rowcolor{white} \multirow{2}{*}{\textbf{Models}} & \multirow{2}{*}{\textbf{Method}} & \multicolumn{6}{c}{\textsc{Executable Evaluation}} & \multicolumn{3}{c}{\textsc{IE Evaluation}} & \multicolumn{2}{c}{\textsc{Exec Eval}} & \multicolumn{1}{c}{\textsc{IE Eval}} & \multicolumn{2}{c}{\textsc{Exec Eval}} & \multicolumn{1}{c}{\textsc{IE Eval}}\\
            
            \cmidrule(r){3-8}\cmidrule(r){9-11}\cmidrule(r){12-13}\cmidrule(r){14-14}\cmidrule(r){15-16}\cmidrule(r){17-17}
            
             \rowcolor{white} & & Correct($\uparrow$) & Prec & Reca & Unex.($\downarrow$) & Over. & Else & Prec & Reca & F1 & Correct & Unex. & F1 & Correct & Unex. & F1 \\
        \midrule
            \multicolumn{17}{c}{\cellcolor{lightgray!50}\textbf{\textit{Closed-source LLMs}}} \\
        \midrule
            \multirow{3}{*}{\parbox{0.15\columnwidth}{GPT-3.5-Turbo}} & COT & 36.75 & 8.83 & 6.71 & 43.46 & 0.71 & 3.53 & 89.45 & 50.43 & 47.99 & 35.19 & 55.40 & 41.28 & 32.65 & 53.06 & 41.16 \\
            & Reflexion & 46.29 & 11.66 & 2.83 & 37.10 & 0.71 & 1.41 & \textbf{94.67} & 55.85 & 55.10 & 43.90 & 49.13 & 48.66 & 36.73 & 51.02 & 43.75 \\
            & \ourmethod & \textbf{54.84} & 11.83 & 8.96 & \textbf{19.35} & 1.08 & 3.94 & 85.85 & \textbf{73.34} & \textbf{69.20} & \textbf{46.34} & \textbf{34.84} & \textbf{57.74} & \textbf{48.98} & \textbf{44.90} & \textbf{52.38} \\

            \noalign{\vskip 0.4mm}
            \hdashline
            \noalign{\vskip 0.4mm}
            \multirow{3}{*}{\parbox{0.15\columnwidth}{Gemini Pro}} & COT & 29.69 & 10.94 & 7.50 & 47.19 & 1.25 & 3.44 & 81.21 & 45.22 & 41.81 & 34.49 & 49.13 & 42.40 & 17.72 & 75.95 & 22.10 \\
            & Reflexion & 33.12 & 6.56 & 4.06 & 52.50 & 0.63 & 3.12 & \textbf{87.45} & 42.75 & 40.88 & 34.15 & 51.57 & 41.66 & 20.25 & 65.82 & 27.66 \\
            & \ourmethod & \textbf{42.81} & 11.87 & 4.69 & \textbf{34.38} & 1.25 & 5.00 & 85.70 & \textbf{57.54} & \textbf{54.91} & \textbf{35.89} & \textbf{42.86} & \textbf{47.80} & \textbf{43.04} & \textbf{34.18} & \textbf{56.92} \\

            \noalign{\vskip 0.4mm}
            \hdashline
            \noalign{\vskip 0.4mm}
            \multirow{3}{*}{\parbox{0.15\columnwidth}{GPT-4-o-mini}} & COT & 54.66 & 13.50 & 6.43 & 20.26 & 0.96 & 4.18 & 89.74 & 72.87 & 69.92 & 45.79 & 38.72 & 56.32 & 46.99 & 42.17 & 53.77 \\
            & Reflexion & 53.70 & 15.11 & 3.22 & 22.83 & 0.96 & 4.18 & \textbf{92.14} & 70.20 & 69.15 & 39.06 & 47.47 & 48.66 & 38.55 & 45.78 & 43.86 \\
            & \ourmethod & \textbf{62.06} & 14.15 & 3.86 & \textbf{15.11} & 0.96 & 3.86 & 91.76 & \textbf{78.10} & \textbf{76.97} & \textbf{56.23} & \textbf{27.27} & \textbf{67.56} & \textbf{53.01} & \textbf{34.94} & \textbf{60.10} \\
            
            \noalign{\vskip 0.4mm}
            \hdashline
            \noalign{\vskip 0.4mm}
            \multirow{3}{*}{\parbox{0.15\columnwidth}{GPT-4-Turbo}} & COT & 61.88 & 12.50 & 7.19 & 14.37 & 0.94 & 3.12 & 87.75 & 79.90 & 76.95 & 56.10 & 29.27 & 65.08 & 50.60 & 30.12 & 64.73 \\
            & Reflexion & 67.50 & 13.75 & 4.37 & 10.94 & 0.94 & 2.50 & \textbf{93.28} & 82.76 & 82.40 & \textbf{64.81} & 19.51 & 75.85 & 50.60 & 33.73 & 63.50 \\
            & \ourmethod & \textbf{71.56} & 14.06 & 5.31 & \textbf{4.06} & 0.63 & 4.37 & 92.49 & \textbf{89.13} & \textbf{88.69} & 64.11 & \textbf{15.33} & \textbf{76.21} & \textbf{57.83} & \textbf{16.87} & \textbf{75.52} \\
        \midrule
            \multicolumn{17}{c}{\cellcolor{lightgray!50}\textbf{\textit{Open-source LLMs}}} \\
        \midrule
            \multirow{3}{*}{\parbox{0.15\columnwidth}{Phi-3-medium}} & COT & 12.50 & 2.81 & 3.12 & 80.00 & 0.00 & 1.56 & \textbf{94.38} & 18.10 & 17.21 & 11.78 & 79.46 & 16.28 & 9.64 & 85.54 & 12.28 \\
            & Reflexion & 12.19 & 6.56 & 1.87 & 77.81 & 0.00 & 1.56 & 92.45 & 18.21 & 17.31 & 12.66 & 82.28 & 15.42 & 7.23 & 90.36 & 8.89 \\
            & \ourmethod & \textbf{24.06} & 12.50 & 7.50 & \textbf{52.19} & 0.31 & 3.44 & 85.07 & \textbf{38.59} & \textbf{34.93} & \textbf{21.15} & \textbf{64.42} & \textbf{30.29} & \textbf{22.89} & \textbf{69.88} & \textbf{26.60} \\
            
            \noalign{\vskip 0.4mm}
            \hdashline
            \noalign{\vskip 0.4mm}
            \multirow{3}{*}{\parbox{0.15\columnwidth}{CodeLlama}} & COT & 17.98 & 3.75 & 2.25 & 74.53 & 0.00 & 1.50 & \textbf{79.75} & 21.98 & 21.36 & 9.01 & 85.84 & 11.21 & 2.70 & 89.19 & 9.19\\
            & Reflexion & 18.08 & 4.80 & 2.95 & 73.06 & 0.00 & 1.11 & 78.96 & 23.26 & 22.44 & \textbf{13.73} & \textbf{80.26} & \textbf{16.01} & 8.82 & 85.29 & 12.69\\
            & \ourmethod & \textbf{23.99} & 8.12 & 1.48 & \textbf{64.94} & 0.00 & 1.48 & 78.59 & \textbf{28.70} & \textbf{28.41} & 11.16 & 85.84 & 12.52 & \textbf{13.51} & \textbf{81.08} & \textbf{17.39} \\
    
            \noalign{\vskip 0.4mm}
            \hdashline
            \noalign{\vskip 0.4mm}
            \multirow{3}{*}{\parbox{0.15\columnwidth}{Mixtral 8$\times$7B}} & COT & 28.75 & 8.13 & 4.37 & 57.81 & 0.31 & 0.63 & \textbf{89.79} & 38.23 & 37.26 & 32.40 & 57.14 & 38.30 & 17.72 & 74.68 & 22.01 \\
            & Reflexion & 36.25 & 6.88 & 3.12 & 51.25 & 0.00 & 2.50 & 89.35 & 44.57 & 43.60 & 29.62 & 62.02 & 33.64 & 22.78 & 69.62 & 28.20 \\
            & \ourmethod & \textbf{46.88} & 10.62 & 7.19 & \textbf{30.31} & 0.63 & 4.37 & 87.32 & \textbf{62.71} & \textbf{59.75} & \textbf{40.77} & \textbf{38.33} & \textbf{52.50} & \textbf{36.71} & \textbf{43.04} & \textbf{48.23}\\
    
            \noalign{\vskip 0.4mm}
            \hdashline
            \noalign{\vskip 0.4mm}
            \multirow{3}{*}{\parbox{0.15\columnwidth}{Deepseek-coder}} & COT & 36.56 & 10.94 & 5.63 & 42.50 & 0.63 & 3.75 & 86.05 & 48.78 & 47.05 & \textbf{38.33} & \textbf{47.74} & \textbf{44.80} & 25.30 & 60.24 & 35.65 \\
            & Reflexion & 37.19 & 11.25 & 4.06 & 44.69 & 1.25 & 1.56 & \textbf{86.41} & 48.28 & 47.08 & 36.24 & 51.92 & 43.64 & 22.89 & 65.06 & 32.04 \\
            & \ourmethod & \textbf{38.75} & 11.25 & 5.31 & \textbf{39.69} & 0.63 & 4.37 & 84.91 & \textbf{52.11} & \textbf{49.68} & 37.63 & 50.52 & 44.33 & \textbf{39.76} & \textbf{42.17} & \textbf{50.28} \\
        \bottomrule
        \end{tabular}
    }
    \caption{The executable evaluation and IE evaluation of LLMs with three frameworks in \textsc{Swde}, \textsc{Extended Swde}, and \textsc{Ds1} dataset. Best Correct, Unexecutable, precision, recall, and F1 score are marked \textbf{bold}.}
    \label{tab:mainexp}
\end{table*}

\subsection{Progressive Generation}
Dealing with the lengthy content and hierarchical structure of webpages, generating a complete and executable scraper in one turn is difficult.
However, the HTML content is organized in a DOM tree structure, which makes it possible to prune irrelevant page components and hence, limit the length and height of the DOM tree to improve the performance of LLM generation.

Specifically, we perform a traversal strategy consisting of \textbf{top-down} and \textbf{step-back} operations. 
\textbf{Top-down} refers to starting from the root node of the current DOM tree, progressively refining down to the specific node containing the target information.
\textbf{Step-back} refers to reassessing and adjusting selection criteria by moving up the DOM tree to choose a more reliable and broadly applicable node as a foundation for more consistent and accurate XPath targeting.
At each step, we first employ a top-down operation, guiding the LLMs to directly write out the XPath leading to the node containing the target information and to judge whether the value extracted with XPath is consistent with the value it recognizes.
If execution fails, then adopt a step-back operation to retreat from the failed node, ensuring the web page includes the target information, which is driven by LLMs.
The detail is shown in Algorithm~\ref{alg:algorithm}.

\subsection{Synthesis}\label{sec:synthesis}

Although we gain an executable action sequence within the progressive generation process, there are still differences in the specific location of the target information and the structure between different web pages. The action sequence may collect XPath with specific characteristics in a single HTML and lose generalizability.
To enhance the reusability of the action sequence, we propose a synthesis phase.

Specifically, we randomly select $n_s$ webpages from the case as seed webpages. Then, we generate an action sequence for each of them. Subsequently, we execute multiple different action sequences to extract information from the seed web pages, respectively. We collect all action sequences and their corresponding results and then choose one that can extract all the target information in the web pages as the final action sequence.
\section{Experiment}
Intending to put \ourmethod to practical use, we investigate the following research questions:
\begin{inparaenum}[1)]
    \item Can \ourmethod outperform the state-of-the-art scraper generation methods?
    \item How does \ourmethod framework improve the performance of the scraper generation task? 
    \item Does \ourmethod meet the requirements for web scraping tasks, specifically being accurate and efficient?
\end{inparaenum}

\subsection{Experimental Settings \& Evaluation Metrics}\label{sec:exp_setting}
We conduct our experiment on 8 LLMs including closed-source LLMs: \textbf{GPT-3.5-Turbo}~\cite{openai2022chatgpt}, \textbf{Gemini Pro}~\cite{geminiteam2023gemini}, \textbf{GPT-4-o-mini}~\cite{openai2024gpt4omini} and \textbf{GPT-4-Turbo}~\cite{openai2023gpt4} as well as open-source LLMs: \textbf{Phi-3-medium}~\cite{abdin2024phi3technicalreporthighly}, \textbf{CodeLlama-34B}~\cite{rozire2024code}, \textbf{Mixtral 8$\times$7B}~\cite{jiang2024mixtral} and \textbf{Deepseek-Coder-33B}~\cite{guo2024deepseekcoder}. 
Furthermore, we apply different LLM-prompt-based web agents as our baselines, including \textbf{COT}~\cite{wei2023chainofthought} and \textbf{Reflexion}~\cite{shinn2023reflexion} and \textbf{\ourmethod}to them. The comparison between them is discussed in Appendix~\ref{apd:baseline}. Due to the limited-length context of LLMs, all experiments are conducted under zero-shot settings. 

We test them on three datasets: \textsc{Swde}~\cite{swde_dataset}, \textsc{Extend Swde}~\cite{lockard2019openceres} and \textsc{Ds1}~\cite{omari2017synthesis}. The detailed experimental results of the last two can be found in Appendix~\ref{apd:extend_swde} and ~\ref{apd:ds1}. We set the size of seed webpages $n_s=3$ for \textsc{Swde} and \textsc{Extend Swde}, $n_s=1$ for \textsc{Ds1} and max retry times $d_{max}=5$.

In addition to the execution evaluation metrics described in Section~\ref{sec:evaluation}, we also employ traditional evaluation metrics to more comprehensively assess the quality of different action sequences. Specifically, we adopt precision (P.), recall (R.), and macro-f1 (F1), which are calculated as the mean of the corresponding metrics for each case. Detailed experimental results on the last two datasets can be found in Table~\ref{tab:dataset_ext_swde} and~\ref{tab:dataset_ds1}.

\subsection{Main Results}
Results in Table~\ref{tab:mainexp} show that:
\begin{inparaenum}[1)]
    \item With \ourmethod generating action sequence, LLMs can achieve better performance. Compared to the COT and Reflexion baseline, our method performs a higher ratio of correct and a lower ratio of un-executable. Also, it should be noted that Mixtral 8$\times$7B + \ourmethod can outperform GPT-3.5-Turbo + Reflexion, indicating the superiority of \ourmethod in the generation of executable action sequences in the scraper generation task.
    \item Models with small parameter sizes have significant difficulties in understanding and writing executable paths, so they can be considered challenging to apply in this task. On the contrary, large-scale models demonstrate a more stable ability in instruction alignment, web structure comprehension, and reflection on execution results.
    \item Traditional IE evaluation metrics cannot well describe the success rate of our task. Especially for the precision metric, it fails to reveal the performance gap among different methods with different models. This is because the extraction metrics only evaluate the results that have been extracted, ignoring that unexecutable or empty extractions also greatly damage the executability.
\end{inparaenum}

\begin{table}
    \centering
    \resizebox{\columnwidth}{!}{\begin{tabular}{llccc}
    \toprule
        \multirow{2}{*}{\textbf{Models}} & \multirow{2}{*}{\textbf{Method}} & \multicolumn{2}{c}{\textsc{Exec Eval}} & \multicolumn{1}{c}{\textsc{IE Eval}} \\
        \cmidrule(r){3-4}\cmidrule(r){5-5}
        & & \textbf{Correct($\uparrow$)} & \textbf{Unex.($\downarrow$)} & \textbf{F1}\\
    \midrule
        \multirow{6}{*}{\parbox{0.2\columnwidth}{GPT-3.5-Turbo}} & COT & 36.75 & 43.46 & 47.99\\
        & \quad \quad - \textit{synthesis} & 27.56 & 57.24 & 34.44\\
        
        \noalign{\smallskip}
        \cdashline{2-5}
        \noalign{\smallskip}
        & Reflexion & 46.29 & 37.10 & 55.10\\
        & \quad \quad - \textit{synthesis} & 28.62 & 59.01 & 35.01 \\

        \noalign{\smallskip}
        \cdashline{2-5}
        \noalign{\smallskip}
        & \ourmethod & 54.84 & 19.35 & 69.20\\
        & \quad \quad - \textit{synthesis} & 44.52 & 29.33 & 58.44\\
    \midrule
        \multirow{6}{*}{\parbox{0.2\columnwidth}{Gemini Pro}} & COT & 29.69 & 47.19 & 41.81 \\
        & \quad \quad - \textit{synthesis} & 27.56 & 57.24 & 33.09\\

        \noalign{\smallskip}
        \cdashline{2-5}
        \noalign{\smallskip}
        & Reflexion & 33.12 & 52.50 & 40.88\\
        & \quad \quad - \textit{synthesis} & 28.62 & 59.01 & 37.60\\

        \noalign{\smallskip}
        \cdashline{2-5}
        \noalign{\smallskip}
        & \ourmethod &  42.81 & 34.38 & 54.91 \\
        & \quad \quad - \textit{synthesis} & 39.46 & 31.56 & 56.48\\
    \midrule
        \multirow{6}{*}{\parbox{0.2\columnwidth}{GPT-4-Turbo}} & COT & 61.88 & 14.37 & 76.95\\
        & \quad \quad - \textit{synthesis} & 46.88 & 30.00 & 61.20\\

        \noalign{\smallskip}
        \cdashline{2-5}
        \noalign{\smallskip}
        & Reflexion & 67.50 & 10.94 & 82.40\\
        & \quad \quad - \textit{synthesis} & 56.87 & 25.31 & 69.78\\

        \noalign{\smallskip}
        \cdashline{2-5}
        \noalign{\smallskip}
        & \ourmethod &  71.56 & 4.06 & 88.69\\
        & \quad \quad - \textit{synthesis} & 65.31 & 11.87 & 80.41\\
    \bottomrule
    \end{tabular}}
    \caption{Ablation study on \textsc{AutoScraper}. We report \textbf{Correct}, \textbf{Unexecutable} from the executive evaluation, and \textbf{F1} score from the IE evaluation in \textsc{Swde} dataset.}
    \label{tab:ablation}
\end{table}
\subsection{Ablation Study}
To further justify the effectiveness of each component of \textsc{AutoScraper}, we perform an ablation study. The results are shown in Table~\ref{tab:ablation}. 
It shows that:
\begin{inparaenum}[1)]
    \item \ourmethod without a second module still beat the other two baseline methods among different LLMs.
    \item The second module of \textsc{AutoScraper}, \textbf{synthesis} module, not only improves \textsc{AutoScraper}, but also improves the performance of other methods. Using more web pages for inference can make the generated scraper more stable and have better generalization.
\end{inparaenum}

\begin{figure}[t]
	\centering
	\includegraphics[width=0.9\columnwidth]{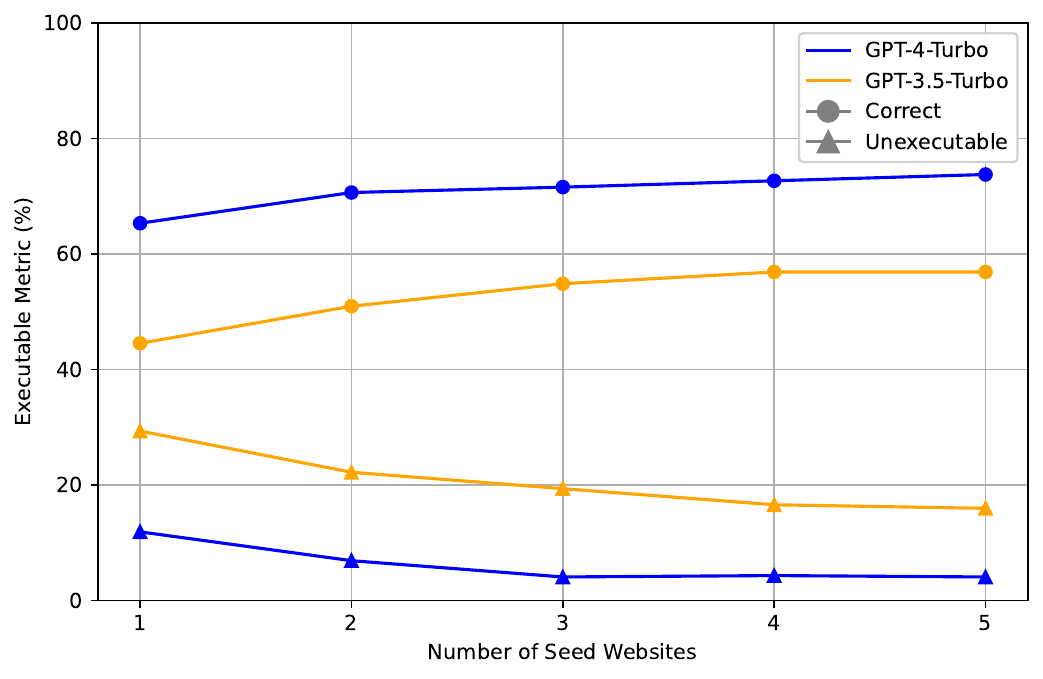}
	\caption{The performance of \ourmethod with different number of seed websites in \textsc{Swde} dataset.}
	\label{fig:seed}
\end{figure}

\subsection{Seed Websites}
In all previous experiments, we fixed the number of seed websites $n_s=3$, which demonstrates the effectiveness of the synthesis module. In this experiment, we offer different numbers of seed webpages and test the performance of \textsc{AutoScraper}. The result is shown in Figure~\ref{fig:seed}.

As the number of seed webpages increases, the correct ratio increases, while the unexecutable ratio decreases. It suggests that the performance of \ourmethod can still be further improved by providing more seed webpages. In addition, the performance improvement reduces as the number increases, which shows that there is an upper limit to improve the performance of \ourmethod by increasing the number of seed webpages.

\section{Discussion}
In this section, we will discuss other aspects of \textsc{AutoScraper}, including its comparison with existing website information extraction methods, efficiency analysis of \textsc{AutoScraper}, and the limitations of the current approach.

\begin{table}[t]
    \centering
    \small
    \begin{tabular}{lcc}
    \toprule
       \textbf{Model} & \textbf{Direct Extraction} & \textbf{\ourmethod} \\
    \midrule
        GPT-3.5-Turbo & 75.76 & 69.20 \\
        Gemini Pro & 76.62 & 54.91 \\
        GPT-4-o-mini & 79.93 & 76.97 \\
        GPT-4-Turbo & 78.56 & 88.69 \\
    \midrule
    \midrule
        Phi-3-medium & 71.73 & 34.93 \\ 
        Codellama & 47.38 & 28.41 \\
        Mixtral 8$\times$7B & 73.45 & 59.75 \\
        Deepseek-coder & 61.96 & 49.68 \\
    \bottomrule
    \end{tabular}
    \caption{Comparing LLM direct extraction with \textsc{AutoScraper} on the \textsc{Swde} dataset.}
    \label{tab:direct}
\end{table}
\subsection{Comparison with LLM Direct Extraction}
Since LLMs can understand human instructions and webpage text, a natural web information extraction solution involves using prompts to guide LLMs to extract target content, which we refer to as direct extraction. We compare direct extraction with \textsc{AutoScraper} both in zero-shot settings using each of the LLMs mentioned above.

Table~\ref{tab:direct} shows that in the direct extraction setting, the extraction performance of all LLMs other than GPT-4-Turbo is superior to that of \textsc{AutoScraper}. However, as the capability of LLMs improves, the gap between the two settings narrows. This indicates that:
\begin{inparaenum}[1.]
    \item While LLMs like Phi-3-medium can understand webpage content well \textit{(i.e., correctly extract the expected content)}, they still struggle to comprehend webpage structures \textit{(i.e., generating XPath using features like DOM tree)}.
    \item \textsc{AutoScraper}, combined with the best current LLMs, already achieves superior extraction performance, and the framework is expected to deliver even better and more stable performance as LLMs continue to improve.
\end{inparaenum}

\begin{table}
    \centering
    \small
    \begin{tabular}{lc}
    \toprule
       \textbf{Model} & \textbf{F1}\\
    \midrule
       Render-Full~\cite{swde_dataset} & 84.30\\
       FreeDOM~\cite{lin2020freedom} & 82.32 \\
       SimpDOM~\cite{zhou2021simplified}  & 83.06\\
       MarkupLM$_{\text{BASE}}$~\cite{li2022markuplm} & 84.31\\
       WebFormer~\cite{wang2022webformer} & 86.58 \\
    \midrule
    \midrule
        Reflexion + GPT-4-Turbo  & 82.40 \\ 
       \ourmethod + GPT-4-Turbo  & \textbf{88.69} \\
    \bottomrule
    \end{tabular}
    \caption{Comparing the extraction performance (F1) of 5 baseline models to our method \ourmethod using GPT-4-Turbo on the \textsc{Swde} dataset. Each value of the supervised model in the table is trained on 1 seed site.}
    \label{tab:supervised}
\end{table}
\subsection{Comparison with supervised baselines}
To further demonstrate that \ourmethod is adaptive to different web information extraction tasks, we conduct a comparison with 5 baseline models in web information extraction on supervised learning scenarios: Render-Full~\cite{swde_dataset} proposes a complicated heuristic algorithm for computing visual distances between predicted value nodes and adjusting the predictions. FreeDOM~\cite{lin2020freedom} and SimpDOM~\cite{zhou2021simplified} encode textual features of DOM tree node with LSTM, while MarkupLM~\cite{li2022markuplm} is pre-trained on HTML with text and markup information jointly. WebFormer~\cite{wang2022webformer} leverages the web layout for effective attention weight computation. These models are trained on webpages in some seed websites and tested on the other websites.

Table \ref{tab:supervised} shows the result. Although the comparison is unfair because our method is in zero-shot settings, \ourmethod beat all of them on F1 scores. It shows that by designing an appropriate framework, LLMs can surpass supervised learning methods in some web information extraction tasks.

\subsection{Efficiency Analysis}
Suppose the number of seed webpages is $n_s$, the number of webpages on the same website is $N_\mathcal{W}$, the time to generate a wrapper is $T_g$, the time of synthesis is $T_s$, and the time for extracting information from a webpage with a wrapper is $T_e$. The total time for extracting all information from all websites with \ourmethod is 
\begin{equation}
    T_1 = T_G + T_E = (n_sT_g + T_s) + N_\mathcal{W}T_e
\end{equation}

Besides, the time for LLMs directly extracting information from a webpage is $T_d$, and the total time for extracting all information from all websites directly is 
\begin{equation}
    T_2 = N_\mathcal{W}T_d
\end{equation}

In a real-world scenario, there are many web pages from the same websites to be extracted. Although generating a wrapper takes more time than extracting directly from a single webpage, the extraction efficiency of subsequent web pages would be significantly improved. To explore how many webpages are needed to make \ourmethod more efficient in web IE, we calculate the threshold of $N_\mathcal{W}$. Suppose $T_1 \le T_2$, we have
\begin{equation}
    T_G + T_E = (n_sT_g + T_s) + N_\mathcal{W}T_e \le N_\mathcal{W}T_d
\end{equation}

\begin{equation}\label{eqa}
    N_\mathcal{W} \ge \frac{n_sT_g + T_s}{T_d - T_e}
\end{equation}

\begin{table}[t]
    \centering
    \small
    \begin{tabular}{lcccc}
    \toprule
       \textbf{Websites} & $T_d$ & $n_sT_g + T_s$ & $T_e$ & $N_\mathcal{W}$ \\
    \midrule
        Auto & 8.27s & 238.4s & 0.30s & 30 \\
        Book & 10.20s & 176.4s & 0.51s & 18 \\
        Camera & 6.59s & 107.1s & 0.31s & 18 \\
        Job & 7.42s & 123.5s & 0.21s & 18 \\
        Movie & 7.47s & 133.2s & 0.21s & 19 \\ 
        Nbaplayer & 8.32s & 179.4s & 0.45s & 23 \\
        Restaurant & 8.87s & 160.8s & 0.54s & 20 \\
        University & 14.26s & 134.7s & 0.32s & 10 \\
    \bottomrule
    \end{tabular}
    \caption{Time efficiency analysis on GPT-4-Turbo.}
    \label{tab:efficiency}
\end{table}

To verify the efficiency advantages of \ourmethod in large-scale web information extraction scenarios, we conducted tests on the \textsc{Swde} dataset. Specifically, we randomly selected a website in each of the 10 domains. 
We repeat 3 times on \ourmethod and record the average time to estimate $n_sT_g + T_s$ and $T_e$. At the same time, we record the average time $T_d$ on 10 web pages with LLM extracting directly. We calculate the threshold of $N_\mathcal{W}$ following the Equation~\ref{eqa} and show them in Table~\ref{tab:efficiency}. It can be observed that the threshold of the page numbers is 19.5 on average, which is significantly lower than the average number of web pages per site in \textsc{Swde} dataset.

\subsection{Error Analysis}
We perform an analysis by looking at the recorded action sequence of \ourmethod with GPT-4-Turbo and identify the following common failure modes. We mainly focus on the cases categorized as unexecutable, over-estimate, and else.

\paragraph{Non-generalizability of webpages} 
The target information and corresponding webpage structures exhibit variations across different webpages, leading to a lack of generalizability in \ourmethod (i.e., the inability to apply the same rules across all webpages in the same website).
For instance, for the task \textit{"Please extract the name of the company offering the job"} in the website job-careerbuilder, most webpages contain the company name, but there is one webpage where the company name is \textit{"Not Available"} on another node of DOM tree.

\paragraph{Miss in multi-valued}
Presented with the task of generating a scraper for extracting \textit{address} in restaurant webpages or \textit{contact phone number} from university websites, the target information is located in multiple locations in the webpage, such as the information bar, title, etc. Although \ourmethod is capable of generating action sequences to extract portions of information, crafting a comprehensive action sequence that captures all of the information remains a challenge.

\section{Conclusion}
In this paper, we introduce the scraper generation task and the paradigm that combines LLMs and scrapers to improve the reusability of the current language-agent-based framework. We then propose \ourmethod, a two-phase framework including progressive generation and synthesis module to generate a more stable and executable action sequence. Our comprehensive experiments demonstrate that \ourmethod can outperform the state-of-the-art baseline in the scraper generation task.

\section*{Acknowledgement}
This work was supported by National Natural Science Foundation of China (No. 62102095).
The computations in this research were performed using the CFFF platform of Fudan University. The authors would like to express their sincere gratitude to Alibaba (China) Co., Ltd. and Alibaba Holding-Aicheng Technology-Enterprise Intelligence Business Unit for their support. 

\section*{Limitation}
We introduce a paradigm that combines LLMs with scrapers for web scraper generation tasks and propose \ourmethod to generate an executable action sequence with progressively understanding the HTML documents. Though experimental results show the effectiveness of our framework, there are still some limits to our work.

First, our framework is restricted to the paradigm in the information extraction task for vertical web pages. LLMs with scrapers provide high efficiency in open-world web IE tasks, but can hardly transfer to existing web environments such as Mind2Web~\cite{deng2023mind2web}, WebArena~\cite{zhou2023webarena}.

Second, our framework relies on the performance of backbone LLMs. Enhancing LLMs' ability to understand HTML is a very valuable research question, including corpus collection and training strategy. We will research HTML understanding enhancement in future work.

\section*{Ethic statement}
We hereby declare that all authors of this article are aware of and adhere to the provided ACL Code of Ethics and honour the code of conduct.

\paragraph{Use of Human Annotations}
Human annotations are only utilized in the early stages of methodological research to assess the feasibility of the proposed solution. All annotators have provided consent for the use of their data for research purposes. We guarantee the security of all annotators throughout the annotation process, and they are justly remunerated according to local standards. Human annotations are not employed during the evaluation of our method.

\paragraph{Risks}
The datasets used in the paper have been obtained from public sources and anonymized to protect against any offensive information. Though we have taken measures to do so, we cannot guarantee that the datasets do not contain any socially harmful or toxic language.

\bibliography{anthology,custom}
\bibliographystyle{acl_natbib}

\newpage
\appendix
\newpage
\label{sec:appendix}
\section{Experiments}

\newcolumntype{o}{>{\columncolor{red!10}}c}
\newcolumntype{b}{>{\columncolor{blue!10}}c}
\renewcommand{\arraystretch}{1.0}
\begin{table*}[h!]
    \centering
    \scriptsize
    \begin{tabular}{lcoooooobbb}
    \toprule
        \multirow{2}{*}{\textbf{Models}} & \multirow{2}{*}{\textbf{Method}} & \multicolumn{6}{c}{\textsc{Executable Evaluation}} & \multicolumn{3}{c}{\textsc{IE Evaluation}}\\
        \cmidrule(r){3-8}\cmidrule(r){9-11}
         \rowcolor{white} & & Correct($\uparrow$) & Prec & Reca & Unex.($\downarrow$) & Over. & Else & Prec & Reca & F1\\
    \midrule
        \multicolumn{11}{c}{\cellcolor{lightgray!50}\textbf{\textit{Closed-source LLMs}}} \\
    \midrule
        \multirow{3}{*}{GPT-3.5-Turbo} & COT & 35.19 & 3.48 & 4.53 & 55.40 & 0.35 & 1.05 & 88.66 & 42.86 & 41.28 \\
        & Reflexion & 43.90 & 1.74 & 2.09 & 49.13 & 0.35 & 2.79 & \textbf{93.46} & 49.58 & 48.66 \\
        & \ourmethod & \textbf{46.34} & 4.18 & 8.01 & \textbf{34.84} & 0.35 & 6.27 & 84.65 & \textbf{61.88} & \textbf{57.74}\\

        \noalign{\vskip 0.4mm}
        \hdashline
        \noalign{\vskip 0.4mm}
        \multirow{3}{*}{Gemini Pro} & COT & 34.49 & 2.09 & 6.62 & 49.13 & 0.35 & 7.32 & 81.09 & 46.55 & 42.40 \\
        & Reflexion & 34.15 & 2.09 & 6.97 & 51.57 & 0.35 & 4.88 & \textbf{84.43} & 45.19 & 41.66\\
        & \ourmethod & \textbf{35.89} & 5.23 & 10.10 & \textbf{42.86} & 0.35 & 5.57 & 83.80 & \textbf{52.83} & \textbf{47.80}\\

        \noalign{\vskip 0.4mm}
        \hdashline
        \noalign{\vskip 0.4mm}
        
        \multirow{3}{*}{GPT-4-o-mini} & COT & 45.79 & 4.38 & 4.71 & 38.72 & 0.00 & 0.64 & 88.59 & 57.97 & 56.32 \\
        & Reflexion & 39.06 & 7.07 & 2.02 & 47.47 & 0.34 & 4.04 & \textbf{95.03} & 49.07 & 48.66 \\
        & \ourmethod & \textbf{56.23} & 5.39 & 5.05 & \textbf{27.27} & 0.00 & 6.06 & 91.12 & \textbf{69.45} & \textbf{67.56} \\
        
        \noalign{\vskip 0.4mm}
        \hdashline
        \noalign{\vskip 0.4mm}
        
        \multirow{3}{*}{GPT-4-Turbo} & COT & 56.10 & 2.44 & 7.32 & 29.27 & 0.35 & 4.53 & 85.15 & 68.35 & 65.08 \\
        & Reflexion & \textbf{64.81} & 4.18 & 5.57 & 19.51 & 0.35 & 5.57 & \textbf{87.39} & 77.81 & 75.85\\
        & \ourmethod & 64.11 & 3.48 & 6.27 & \textbf{15.33} & 0.35 & 10.45 & 82.71 & \textbf{80.25} & \textbf{76.21} \\
    \midrule
        \multicolumn{11}{c}{\cellcolor{lightgray!50}\textbf{\textit{Open-source LLMs}}} \\
    \midrule
        \multirow{3}{*}{Phi-3-medium} & COT & 11.78 & 1.01 & 5.05 & 79.46 & 0.34 & 2.36 & 91.03 & 19.08 & 16.28 \\
        & Reflexion & 12.66 & 1.90 & 1.90 & 82.28 & 0.00 & 1.27 & \textbf{93.87} & 16.03 & 15.42 \\
        & \ourmethod & \textbf{21.15} & 2.88 & 7.69 & \textbf{64.42} & 0.00 & 3.85 & 87.88 & \textbf{33.39} & \textbf{30.29} \\
        
        \noalign{\vskip 0.4mm}
        \hdashline
        \noalign{\vskip 0.4mm}
        \multirow{3}{*}{CodeLlama} & COT & 9.01 & 1.29 & 2.15 & 85.84 & 0.00 & 1.72 & 87.22 & 12.62 & 11.21 \\
        & Reflexion & \textbf{13.73} & 1.72 & 3.00 & \textbf{80.26} & 0.00 & 1.29 & 89.41 & \textbf{17.76} & \textbf{16.01} \\
        & \ourmethod & 11.16 & 0.00 & 1.72 & 85.84 & 0.00 & 1.29 & \textbf{92.49} & 13.29 & 12.52 \\

        \noalign{\vskip 0.4mm}
        \hdashline
        \noalign{\vskip 0.4mm}
        \multirow{3}{*}{Mixtral 8$\times$7B} & COT & 32.40 & 1.05 & 4.88 & 57.14 & 0.35 & 4.18 & \textbf{87.87} & 41.20 & 38.30\\
        & Reflexion & 29.62 & 1.05 & 4.18 & 62.02 & 0.35 & 2.79 & 83.44 & 36.44 & 33.64\\
        & \ourmethod & \textbf{40.77} & 3.83 & 9.76 & \textbf{38.33} & 0.35 & 6.97 & 82.50 & \textbf{58.14} & \textbf{52.50}\\

        \noalign{\vskip 0.4mm}
        \hdashline
        \noalign{\vskip 0.4mm}
        \multirow{3}{*}{Deepseek-coder} & COT & \textbf{38.33} & 3.83 & 6.62 & \textbf{47.74} & 0.35 & 3.14 & 81.32 & \textbf{48.52} & \textbf{44.80} \\
        & Reflexion & 36.24 & 3.48 & 3.83 & 51.92 & 0.00 & 4.53 & 83.53 & 45.03 & 43.64\\
        & \ourmethod & 37.63 & 2.44 & 5.92 & 50.52 & 0.35 & 3.14 & \textbf{86.91} & 47.09 & 44.33 \\
    \bottomrule
    \end{tabular}
    \caption{The executable evaluation and IE evaluation of LLMs with three frameworks in \textsc{Extended Swde} dataset. We examine 6 LLMs, including 3 closed-source LLMs and 3 open-source LLMs. }
    \label{tab:mainexp_ex_swde}
\end{table*}
\newcolumntype{o}{>{\columncolor{red!10}}c}
\newcolumntype{b}{>{\columncolor{blue!10}}c}

\begin{table*}[t]
    \centering
    \scriptsize
    \begin{tabular}{lcoooooobbb}
    \toprule
        \multirow{2}{*}{\textbf{Models}} & \multirow{2}{*}{\textbf{Method}} & \multicolumn{6}{c}{\textsc{Executable Evaluation}} & \multicolumn{3}{c}{\textsc{IE Evaluation}}\\
        \cmidrule(r){3-8}\cmidrule(r){9-11}
         \rowcolor{white} & & Correct($\uparrow$) & Prec & Reca & Unex.($\downarrow$) & Over. & Else & Prec & Reca & F1\\
    \midrule
        \multicolumn{11}{c}{\cellcolor{lightgray!50}\textbf{\textit{Closed-source LLMs}}} \\
    \midrule
        \multirow{3}{*}{GPT-3.5-Turbo} & COT & 32.65 & 4.08 & 8.16 & 53.06 & 0.00 & 2.04 & 90.56 & 43.54 & 41.16 \\
        & Reflexion & 36.73 & 8.16 & 4.08 & 51.02 & 0.00 & 0.00 & \textbf{95.56} & 44.22 & 43.75 \\
        & \ourmethod & \textbf{48.98} & 4.08 & 0.00 & \textbf{44.90} & 0.00 & 2.04 & 94.90 & \textbf{51.70} & \textbf{52.38}\\

        \noalign{\vskip 0.4mm}
        \hdashline
        \noalign{\vskip 0.4mm}
        \multirow{3}{*}{Gemini Pro} & COT & 17.72 & 2.53 & 3.80 & 75.95 & 0.00 & 0.00 & 90.82 & 22.88 & 22.10 \\
        & Reflexion & 20.25 & 10.13 & 1.27 & 65.82 & 0.00 & 2.53 & 88.83 & 26.93 & 27.66\\
        & \ourmethod & \textbf{43.04} & 15.19 & 3.80 & \textbf{34.18} & 0.00 & 3.80 & \textbf{93.76} & \textbf{55.97} & \textbf{56.92}\\

        \noalign{\vskip 0.4mm}
        \hdashline
        \noalign{\vskip 0.4mm}
        \multirow{3}{*}{GPT-4-o-mini} & COT & 46.99 & 3.61 & 4.82 & 42.17 & 0.00 & 2.41 & 79.74 & 55.34 & 53.77 \\
        & Reflexion & 38.55 & 13.25 & 2.41 & 45.78 & 0.00 & 0.00 & \textbf{91.40} & 43.68 & 43.86 \\
        & \ourmethod & \textbf{53.01} & 6.02 & 4.82 & \textbf{34.94} & 0.00 & 1.20 & 79.06 & \textbf{61.03} & \textbf{60.10}\\
        
        \noalign{\vskip 0.4mm}
        \hdashline
        \noalign{\vskip 0.4mm}
        
        \multirow{3}{*}{GPT-4-Turbo} & COT & 50.60 & 9.64 & 6.02 & 30.12 & 0.00 & 3.61 & 93.60 & 65.75 & 64.73 \\
        & Reflexion & 50.60 & 10.84 & 4.82 & 33.73 & 0.00 & 0.00 & \textbf{96.85} & 62.65 & 63.50\\
        & \ourmethod & \textbf{57.83} & 15.66 & 4.82 & \textbf{16.87} & 0.00 & 4.82 & 92.88 & \textbf{74.95} & \textbf{75.52} \\
    \midrule
        \multicolumn{11}{c}{\cellcolor{lightgray!50}\textbf{\textit{Open-source LLMs}}} \\
    \midrule
        \multirow{3}{*}{Phi-3-medium} & COT & 9.64 & 4.82 & 0.00 & 85.54 & 0.00 & 0.00 & 95.18 & 11.76 & 12.28 \\
        & Reflexion & 7.23 & 0.00 & 1.20 & 90.36 & 0.00 & 1.20 & \textbf{97.87} & 9.47 & 8.89 \\
        & \ourmethod & \textbf{22.89} & 3.61 & 3.61 & \textbf{69.88} & 0.00 & 0.00 & 88.00 & \textbf{28.22} & \textbf{26.60} \\
        
        \noalign{\vskip 0.4mm}
        \hdashline
        \noalign{\vskip 0.4mm}
        \multirow{3}{*}{CodeLlama} & COT & 2.70 & 2.70 & 5.41 & 89.19 & 0.00 & 0.00 & 78.72 & 10.62 & 9.19 \\
        & Reflexion & 8.82 & 0.00 & 5.88 & 85.29 & 0.00 & 0.00 & \textbf{94.12} & 14.41 & 12.69 \\
        & \ourmethod & \textbf{13.51} & 0.00 & 5.41 & \textbf{81.08} & 0.00 & 0.00 & 84.12 & \textbf{18.92} & \textbf{17.39} \\

        \noalign{\vskip 0.4mm}
        \hdashline
        \noalign{\vskip 0.4mm}
        \multirow{3}{*}{Mixtral 8$\times$7B} & COT & 17.72 & 6.33 & 0.00 & 74.68 & 0.00 & 1.27 & \textbf{94.81} & 21.15 & 22.01\\
        & Reflexion & 22.78 & 6.33 & 1.27 & 69.62 & 0.00 & 0.00 & 94.15 & 28.03 & 28.20\\
        & \ourmethod & \textbf{36.71} & 11.39 & 6.33 & \textbf{43.04} & 0.00 & 2.53 & 91.59 & \textbf{48.52} & \textbf{48.23}\\

        \noalign{\vskip 0.4mm}
        \hdashline
        \noalign{\vskip 0.4mm}
        \multirow{3}{*}{Deepseek-coder} & COT & 25.30 & 9.64 & 2.41 & 60.24 & 0.00 & 2.41 & \textbf{92.47} & 34.71 & 35.65 \\
        & Reflexion & 22.89 & 6.02 & 3.61 & 65.06 & 0.00 & 2.41 & 90.21 & 31.43 & 32.04\\
        & \ourmethod & \textbf{39.76} & 10.84 & 6.02 & \textbf{42.17} & 0.00 & 1.20 & 90.43 & \textbf{51.39} & \textbf{50.28} \\
    \bottomrule
    \end{tabular}
    \caption{The executable evaluation and IE evaluation of LLMs with three frameworks in \textsc{Ds1} dataset. We examine 8 LLMs, including 4 closed-source LLMs and 4 open-source LLMs.}
    \label{tab:mainexp_ds1}
\end{table*}

\subsection{Main results on \textsc{Extended Swde}}\label{apd:extend_swde}
Because \textsc{Extended Swde} dataset focuses on \textit{OpenIE} task (the relation is also expected to be extracted), we first 
map relations into a predefined list of attributes and remove unusual ones. Specifically, we conducted experiments with 294 attributes from 21 websites selected from the \textsc{Extended Swde} dataset.

Table~\ref{tab:mainexp_ex_swde} shows the result. By comparing Table~\ref{tab:mainexp}, we find that:
\begin{inparaenum}[1)]
    \item Under complex extraction task settings (multiple target values and ambiguous problem description), the closed-source LLMs perform better in generating executable action sequences compared to the open-source LLMs.
    \item There are some tasks with unclear descriptions, such as the \textit{"Calendar System"} and \textit{"Facilities and Programs Offered"} on university websites, which affect the wrapper generation performance of all methods.
\end{inparaenum}

\subsection{Main results on \textsc{Ds1}}\label{apd:ds1}
Due to \textsc{Ds1} only contains 166 hand-crafted webpages, and for each website, there are only two webpages, so we take one webpage for inference and the other for evaluation. Meanwhile, due to the number of seed websites equal to one, we test three methods without applying the synthesis module described in Section~\ref{sec:synthesis}.

Table~\ref{tab:mainexp_ds1} shows the result in the DS1 dataset.
Among all LLMs with three methods, GPT-4-Turbo + \ourmethod achieves the best performance, and \ourmethod beats the other two methods in all LLMs, which is consistent with our conclusion.

\subsection{Generate with Golden Label}
\renewcommand{\arraystretch}{0.95}
\begin{table}[t]
    \centering
    \resizebox{\columnwidth}{!}{\begin{tabular}{lccccccc}
    \toprule
        \multirow{2}{*}{\textbf{Models}} & \multirow{2}{*}{\textbf{Method}} & \multicolumn{6}{c}{\textsc{Executable Evaluation}}\\
    \cmidrule(r){3-8}
         & & Correct($\uparrow$) & Prec & Reca & Unex.($\downarrow$) & Over. & Else\\
    \midrule
        \multicolumn{8}{c}{\cellcolor{lightgray!50}\textbf{\textit{Closed-source LLMs}}} \\
    \midrule
        \multirow{3}{*}{\parbox{0.2\columnwidth}{GPT-3.5-Turbo}} & COT & 41.70 & 12.92 & 7.38 & 35.42 & 0.74 & 1.85\\
        & Reflexion & 47.23 & 16.24 & 2.21 & 33.21 & 0.37 & 0.74\\
        & \ourmethod & \textbf{56.89} & 19.43 & 5.65 & \textbf{13.43} & 0.71 & 3.89\\
        
        \noalign{\vskip 0.4mm}
        \hdashline
        \noalign{\vskip 0.4mm}
        \multirow{3}{*}{\parbox{0.2\columnwidth}{Gemini Pro}} & COT & 33.44 & 9.38 & 9.06 & 44.69 & 0.94 & 2.50\\
        & Reflexion & 35.31 & 9.38 & 6.88 & 43.75 & 1.56 & 3.12\\
        & \ourmethod & \textbf{45.31} & 13.44 & 6.25 & \textbf{30.31} & 1.25 & 3.44\\

        \noalign{\vskip 0.4mm}
        \hdashline
        \noalign{\vskip 0.4mm}
        \multirow{3}{*}{\parbox{0.2\columnwidth}{GPT-4-o-mini}} & COT & 56.59 & 12.54 & 8.04 & 17.36 & 0.96 & 0.45\\
        & Reflexion & 62.38 & 10.29 & 1.61 & 23.15 & 0.64 & 1.93\\
        & \ourmethod & \textbf{67.20} & 12.22 & 3.86 & \textbf{12.22} & 0.96 & 3.54\\
        
        \noalign{\vskip 0.4mm}
        \hdashline
        \noalign{\vskip 0.4mm}
        \multirow{3}{*}{\parbox{0.2\columnwidth}{GPT-4-Turbo}} & COT & 61.88 & 11.56 & 9.06 & 11.56 & 1.25 & 4.69\\
        & Reflexion & 71.25 & 7.19 & 4.69 & 14.37 & 0.94 & 1.56\\
        & \ourmethod & \textbf{75.31} & 10.94 & 4.37 & \textbf{4.06} & 0.63 & 4.69\\
    \midrule
        \multicolumn{8}{c}{\cellcolor{lightgray!50}\textbf{\textit{Open-source LLMs}}} \\
    \midrule
        \multirow{3}{*}{Phi-3-medium} & COT & 11.11 & 4.13 & 1.27 & 82.22 & 0.00 & 1.27 \\
        & Reflexion & 12.19 & 5.27 & 7.59 & 72.43 & 0.31 & 2.21 \\
        & \ourmethod & \textbf{27.27} & 16.45 & 9.52 & \textbf{41.56} & 0.87 & 4.33  \\
        
        \noalign{\vskip 0.4mm}
        \hdashline
        \noalign{\vskip 0.4mm}
        \multirow{3}{*}{\parbox{0.2\columnwidth}{CodeLlama}} & COT & 21.40 & 6.27 & 2.21 & 66.79 & 0.74 & 2.58\\
        & Reflexion & 22.21 & 4.93 & 3.94 & 66.95 & 0.49 & 1.48 \\
        & \ourmethod & \textbf{26.20} & 12.55 & 5.54 & \textbf{53.51} & 0.00 & 2.21\\

        \noalign{\vskip 0.4mm}
        \hdashline
        \noalign{\vskip 0.4mm}
        \multirow{3}{*}{\parbox{0.2\columnwidth}{Mixtral 8$\times$7B}} & COT & 27.50 & 7.50 & 5.31 & 56.87 & 0.94 & 1.87 \\
        & Reflexion & 34.69 & 8.13 & 5.31 & 49.06 & 0.63 & 2.19 \\
        & \ourmethod & \textbf{45.62} & 11.56 & 5.94 & \textbf{32.50} & 1.25 & 3.12 \\

        \noalign{\vskip 0.4mm}
        \hdashline
        \noalign{\vskip 0.4mm}
        \multirow{3}{*}{\parbox{0.2\columnwidth}{Deepseek-coder}} & COT & 35.00 & 18.75 & 5.31 & 36.25 & 0.63 & 4.06\\
        & Reflexion & \textbf{38.75} & 11.87 & 2.81 & 42.19 & 0.63 & 3.75\\
        & \ourmethod & 38.44 & 20.94 & 4.06 & \textbf{31.56} & 0.94 & 6.56 \\
    \bottomrule
    \end{tabular}}
    \caption{The executable and IE evaluation with 8 LLMs on \textsc{Swde} dataset with golden label.}
    \label{tab:gold_label}
\end{table}
To better illustrate the effectiveness of our framework in generating executable action sequences, we compare the performance of COT, Reflexion, and \ourmethod, while answering the instruction. 
By offering the same extraction targets, we can effectively detect the performance of different frameworks in generating action sequences.

Table~\ref{tab:gold_label} shows experimental results, from which we can have the following observations: 
\begin{inparaenum}[1)]
    \item Our proposed progressive understanding framework still effectively enhances the model's performance under this setting;
    \item LLMs still suffer in accurately understanding web page contents with semi-structured markup languages, which illustrate the performance gap between Table~\ref{tab:mainexp} and Table~\ref{tab:gold_label};
    \item Compared to closed-source LLMs, even provided with golden labels, Open-source LLMs are unable to achieve sustained performance improvement.
    This phenomenon demonstrates that the bottleneck for these models lies not in understanding the webpage content but in understanding the webpage's hierarchical structure itself.
\end{inparaenum}

\SetCommentSty{mycommfont}

\newcommand\mycommfont[1]{\footnotesize\textcolor{gray!70}{\textit{#1}}}

\begin{algorithm}[t]
\caption{Algorithm for progressive understanding}\label{alg:algorithm}

\KwData{origin HTML code $h_0$, task instruction $I$, max retry times $d_{max}$}

\KwResult{Executable action sequence $\mathcal{A}_{seq}$ to extract the value in the HTML}

Initial history $\mathcal{A}_{seq}\leftarrow [], k = 0$;

\While{True}{
    \lIf{$k>d_{max}$}{break}

    \tcp{Top-down}
    $value, xpath \leftarrow \textbf{LLM}_{g}(h_k, I)$\;
    $result \leftarrow \textbf{Parser}_{text}(h_k, xpath)$\;
    \lIf{result == value}{break}
        
    \tcp{Step-back}
    \Repeat{h contains value}{
        $xpath\leftarrow xpath + "/.."$\;
        $h_{k+1} \leftarrow \textbf{Parser}_{node}(h_k, xpath)$\;
    }
    
    $\text{Append}(\mathcal{A}_{seq}, xpath)$\;
    $k \leftarrow k + 1$\;
}

\Return{$\mathcal{A}_{seq}$}
\end{algorithm}

\section{Analysis on \ourmethod}

\subsection{Comparison with COT \& Reflexion}\label{apd:baseline}
\begin{figure*}[t]
    \centering
    \includegraphics[width=\linewidth]{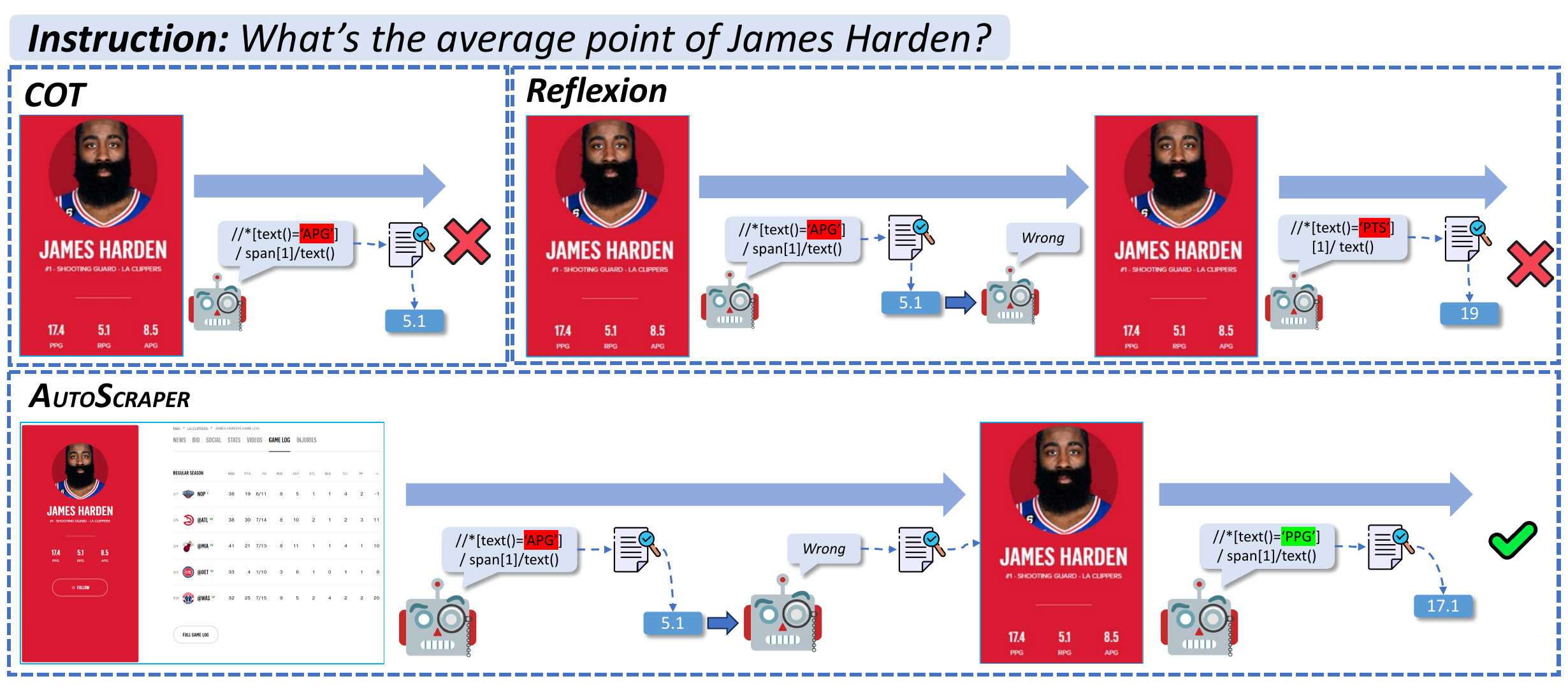}
    \caption{Comparison of \ourmethod with COT and Reflexion.}
    \label{fig:baseline_compare}
\end{figure*}

Figure~\ref{fig:baseline_compare} more intuitively shows the specific differences between different baselines in the experiment.
The most significant difference between \ourmethod and other methods lies in whether the hierarchical structure of web pages is utilized to help LLMs reduce the difficulty of complex web structures. COT only executes one turn while the other executes multiple turns and can learn from the failed execution of the wrapper. Compared to the Reflexion method, \ourmethod employs top-down and step-back operations to prune the DOM tree during each XPath generation process, thereby reducing the length of the web page. In contrast, the Reflexion method can only reflect and regenerate after producing an unexecutable XPath, which does not effectively simplify the webpage. 

\begin{table}
    \centering
    \small
    \begin{tabular}{lcccccc}
    \toprule
        \textbf{Models} & \textbf{1} & \textbf{2} & \textbf{3} & \textbf{4} & \textbf{5} & \textbf{Avg.}\\
    \midrule
        GPT-4-Turbo & 214 & 61 & 13 & 18 & 10 & 1.57\\
        GPT-4-o-mini & 183 & 35 & 20 & 10 & 60 & 2.12\\
        GPT-3.5-Turbo & 124 & 61 & 38 & 22 & 73 & 2.56\\
        Gemini Pro & 94 & 52 & 33 & 27 & 105 & 2.99 \\
        Mixtral 8$\times$7B & 89 & 53 & 43 & 24 & 104 & 3.00 \\
        Phi-3-medium & 47 & 52 & 28 & 26 & 155 & 3.62 \\
        Deepseek-coder & 137 & 70 & 55 & 29 & 23 & 2.14 \\
        CodeLlama & 75 & 35 & 32 & 18 & 80 & 2.97\\
    \bottomrule
    \end{tabular}
    \caption{Length of action sequence of \ourmethod based on different LLMs in \textsc{Swde} dataset.}
    \label{tab:step1}
\end{table}

\begin{table}
    \centering
    \small
    \begin{tabular}{lcccccc}
    \toprule
        \textbf{Models} & \textbf{1} & \textbf{2} & \textbf{3} & \textbf{4} & \textbf{5} & \textbf{Avg.}\\
    \midrule
        GPT-4-Turbo & 28 & 23 & 15 & 11 & 5 & 2.29\\
        GPT-4-o-mini & 50 & 10 & 5 & 1 & 16 & 2.06\\
        GPT-3.5-Turbo & 15 & 10 & 3 & 5 & 7 & 2.48\\
        Gemini Pro & 22 & 17 & 13 & 7 & 20 & 2.82\\
        Mixtral 8$\times$7B & 16 & 13 & 7 & 11 & 29 & 3.32\\
        Phi-3-medium & 14 & 15 & 6 & 2 & 46 & 3.61\\
        Deepseek-coder & 34 & 20 & 17 & 10 & 2 & 2.11\\
        CodeLlama & 18 & 6 & 6 & 9 & 33 & 3.46 \\
    \bottomrule
    \end{tabular}
    \caption{Length of action sequence of \ourmethod based on different LLMs in \textsc{Ds1} dataset.}
    \label{tab:step2}
\end{table}

\begin{table}
    \centering
    \small
    \begin{tabular}{lcccccc}
    \toprule
        \textbf{Models} & \textbf{1} & \textbf{2} & \textbf{3} & \textbf{4} & \textbf{5} & \textbf{Avg.}\\
    \midrule
        GPT-4-Turbo & 61 & 40 & 45 & 53 & 76 & 3.15 \\
        GPT-4-o-mini & 133 & 31 & 17 & 15 & 91 & 2.65 \\
        GPT-3.5-Turbo & 88 & 35 & 48 & 23 & 97 & 3.02 \\
        Gemini Pro & 60 & 41 & 29 & 28 & 132 & 3.45 \\
        Mixtral 8$\times$7B & 51 & 38 & 26 & 29 & 138 & 3.59\\
        Phi-3-medium & 43 & 39 & 34 & 25 & 144 & 3.66 \\
        Deepseek-coder & 120 & 79 & 35 & 33 & 20 & 2.14 \\
        CodeLlama & 53 & 31 & 6 & 6 & 14 & 2.06 \\
    \bottomrule
    \end{tabular}
    \caption{Length of action sequence of \ourmethod based on different LLMs in \textsc{Extended Swde} dataset.}
    \label{tab:step3}
\end{table}
\subsection{Further Study with \ourmethod}
\paragraph{The length of the action sequence is dependent on the LLM capability.}
To comprehensively explore the performance of different LLMs in understanding web page structure, we explore the impact of models on the number distribution of the steps.
In particular, we collect all the action sequences and calculate the average steps of \ourmethod with different LLMs.
The experimental result is reported in Table~\ref{tab:step1},~\ref{tab:step2} and~\ref{tab:step3}.

We observe that \ourmethod with stronger LLMs generates fewer lengths of action sequence. \ourmethod with GPT-4-Turbo generates 1.57 steps on average, while the \ourmethod with Phi-3-medium generates 3.62 steps on average. 
This phenomenon can be interpreted as more powerful models having a better understanding of the web page hierarchical structure, thus being able to accurately output the appropriate XPaths in longer/deeper web pages, thereby reducing the number of steps.





\begin{table}[t]
    \centering
    \small
    {\begin{tabular}{lrr}
    \toprule
        \textbf{Models} & Contains & Equal(=) \\
    \midrule
        GPT4 & 0.61\% & 2.90\% \\
        GPT-3.5-Turbo & 9.33\% & 9.78\% \\
        Gemini Pro & 10.62\% & 14.29\% \\ 
        Mixtral 8$\times$7B & 12.88\% & 8.55\% \\ 
        Deepseek-Coder & 11.63\% & 7.55\% \\ 
        CodeLlama & 18.75\% & 14.29\% \\
        Mistral 7B & 18.18\% & 33.33\% \\ 
    \bottomrule
    \end{tabular}}
    \caption{Bad case ratio in two types of predicate.}
    \label{tab:contains}
\end{table}
\newcommand{\mycolorbox}[2][m_ans]{%
  \colorbox{#1}{\raisebox{0pt}[0.8\height][0.4\depth]{#2}}%
}

\begin{table*}
    \centering
    \small
    \begin{tabularx}{\textwidth}{lXX}
    \toprule
        & \textbf{Good case} & \textbf{Bad case}\\
    \midrule
        \textbf{Question} & \textit{Here's a webpage on detail information with detail information of an NBA player. Please extract the \textbf{height} of the player.} & \textit{Here's a webpage with detailed information about a university. Please extract the \textbf{contact phone number} of the university.}\\
        \midrule
         \textbf{Case} & \textit{//div[@class=`gray200B-dyContent`]/ b[\mycolorbox[green]{\textbf{contains(text(),`Height:`)}}]/following-sibling::text()} & \textit{//div[@class=`infopage`]//h5[\mycolorbox[red]{\textbf{contains}} \mycolorbox[red]{\textbf{(text(), `703-528-7809`)}}]} \\
    \bottomrule
    \end{tabularx}
    \caption{Examples of XPath fragility. The \mycolorbox[green]{green} focuses on the common information across different webpages, while the \mycolorbox[red]{red} focuses on specific information of seed webpages.}
    \label{tab:contains_xmp}
\end{table*}
\paragraph{XPath fragility within \ourmethod} 
The fragility of XPath often refers to the characteristic of XPath expressions becoming ineffective or inaccurately matching the target element when faced with new web pages. This is mainly due to XPath specifying specific information through \textit{predicates}, such as text, \texttt{@class}, etc. 

We mainly focus on the fragility of text because these webpages are from the same websites (i.e. \texttt{@class} is a good characteristic for generating stable action sequences). 
Table~\ref{tab:contains_xmp} shows XPath expressions that rely on text. We aim to explore the reusability of generating XPath based on text features.
We manually calculated the proportion of bad cases with two types of predicates, \textit{contains} and \textit{equal}~\footnote{\url{https://www.w3schools.com/xml/xpath_syntax.asp}}. The results in Table~\ref{tab:contains} show that the stronger LLMs capability, the lower the proportion of bad cases with \ourmethod. However, it should be noted that the current SoTA LLM GPT-4-Turbo still suffers from an XPath fragility problem, which indicates that relying entirely on LLMs to generate reliable XPath still has some distance to go.

\section{Dataset Statistic}\label{apd:sec:dataset}
Table~\ref{apd:dataset_statistic},~\ref{tab:dataset_ext_swde},~\ref{tab:dataset_ds1} shows the detailed statistic about the semi-structure web information extraction dataset \textsc{Swde}, \textsc{Extended Swde} and \textsc{Ds1}.

\section{Prompt List}\label{apd:prompt}\label{apd:task_prompt}

\subsection{Task Prompt}\label{apd:task_prompt}
Table~\ref{tab:task_prompt} shows the task prompt we design for each attribute for \textsc{Swde}.

\subsection{Module Prompt}\label{apd:module_prompt}
We provide a comprehensive list of all the prompts that have been used in this study,
offering a clear reference to understand our experimental approach.

\begin{table*}
    \centering
    \small
    \begin{tabular}{lp{0.13\textwidth}p{0.13\textwidth}p{0.05\textwidth}|lp{0.13\textwidth}p{0.13\textwidth}p{0.05\textwidth}}
    \toprule
        \textbf{Domain} & \textbf{Attribute} & \textbf{Website} & \textbf{Num} & \textbf{Domain} & \textbf{Attribute} & \textbf{Website} & \textbf{Num}\\
    \midrule
        \multirow{10}{*}{\textbf{Auto}} & \multirow{10}{*}{\parbox{0.15\textwidth}{model \\ price \\engine \\fuel\_economy}} & aol & 2000 & \multirow{10}{*}{\textbf{Movie}} & \multirow{10}{*}{\parbox{0.3\textwidth}{ttitle\\director\\genre\\mpaa\_rating}} & allmovie & 2000\\
        & & autobytel & 2000 & & & amctv & 2000 \\
        & & automotive & 1999 & & & boxofficemojo & 2000 \\ 
        & & autoweb & 2000 & & & hollywood & 2000 \\ 
        & & carquotes & 2000 & & & iheartmovies & 2000 \\ 
        & & cars & 657 & & & imdb & 2000 \\ 
        & & kbb & 2000 & & & metacritic & 2000 \\ 
        & & motortrend & 1267 & & & msn & 2000 \\ 
        & & msn & 2000 & & & rottentomatoes & 2000 \\ 
        & & yahoo & 2000 & & & yahoo & 2000 \\
    \midrule

        \multirow{10}{*}{\textbf{Book}} & \multirow{10}{*}{\parbox{0.3\textwidth}{title\\author\\isbn\_13\\publisher\\pub\_date}} & abebooks & 2000 & \multirow{10}{*}{\textbf{NBAPlayer}} & \multirow{10}{*}{\parbox{0.3\textwidth}{name\\team\\height\\weight}} & espn & 434\\
        & & amazon & 2000 & & & fanhouse & 446\\
        & & barnesandnoble & 2000 & && foxsports & 425 \\ 
        & & bookdepository & 2000 & && msnca & 434 \\ 
        & & booksamillion & 2000 & && nba & 434 \\ 
        & & bookorders & 2000 & && si & 515 \\ 
        & & buy & 2000 & && slam & 423 \\ 
        & & christianbook & 2000 & && usatoday & 436 \\ 
        & & deepdiscount & 2000 & && wiki & 420 \\ 
        & & waterstone & 2000 & && yahoo & 438 \\

    \midrule
    
        \multirow{10}{*}{\textbf{Camera}} & \multirow{10}{*}{\parbox{0.3\textwidth}{model\\price\\manufacturer}} & amazon & 1767 & \multirow{10}{*}{\textbf{Restaurant}} & \multirow{10}{*}{\parbox{0.3\textwidth}{name\\address\\phone\\cuisine}} & fodors & 2000\\
        & & beachaudio & 247 & & & frommers & 2000 \\ 
        & & buy & 500 & & & zagat & 2000 \\
        & & compsource & 430 & & & gayot & 2000 \\ 
        & & ecost & 923 & & & opentable & 2000 \\ 
        & & jr & 367 & & & pickaretaurant & 2000 \\ 
        & & newegg & 220 & & & restaurantica & 2000 \\ 
        & & onsale & 261 & & & tripadvisor & 2000 \\ 
        & & pcnation & 234 & & & urbanspoon & 2000 \\ 
        & & thenerd & 309 & & & usdiners & 2000 \\ 
    \midrule
    
        \multirow{10}{*}{\textbf{Job}} & \multirow{10}{*}{\parbox{0.3\textwidth}{title\\company\\location\\date\_posted}} & careerbuilder & 2000 & \multirow{10}{*}{\textbf{University}} & \multirow{10}{*}{\parbox{0.3\textwidth}{name\\phone\\website\\type}} & collegeboard & 2000\\
        & & dice & 2000 & & & collegenavigator & 2000 \\ 
        & & hotjobs & 2000 & & & collegeprowler & 2000 \\ 
        & & job & 2000 & & & collegetoolkit & 2000 \\ 
        & & jobcircle & 2000 & & & ecampustours & 1063 \\ 
        & & jobtarget & 2000 & & & embark & 2000 \\ 
        & & monster & 2000 & & & matchcollege & 2000 \\ 
        & & nettemps & 2000 & & & princetonreview & 615 \\
        & & rightitjobs & 2000 & & & studentaid & 2000 \\ 
        & & techcentric & 2000 & & & usnews & 1027 \\ 
    \bottomrule
    \end{tabular}
    \caption{Detail statistic of \textsc{Swde} dataset.}
    \label{apd:dataset_statistic}
\end{table*}
\begin{table}
    \centering
    \small
    \begin{tabular}{llr}
    \toprule
        \textbf{Domain} & \textbf{Website} & \textbf{\# Attributes}\\
    \midrule
        \multirow{8}{*}{Movie} & allmovie & 20\\
        & amctv & 13\\
        & hollywood & 12\\
        & iheartmovies & 8 \\
        & imdb & 34\\
        & metacritic & 17 \\
        & rottentomatoes & 10 \\
        & yahoo & 10 \\

        \midrule
        \multirow{8}{*}{NBAPlayer} & espn & 10\\
        & fanhouse & 14\\
        & foxsports & 10\\
        & msnca & 12 \\
        & si & 12\\
        & slam & 12 \\
        & usatoday & 5 \\
        & yahoo & 9 \\

        \midrule
        \multirow{5}{*}{University} & collegeprowler & 18\\
        & ecampustours & 14 \\
        & embark & 23\\
        & matchcollege & 15 \\
        & usnews & 19\\
    \bottomrule
    \end{tabular}
    \caption{Detail statistic of \textsc{Extend Swde} dataset.}
    \label{tab:dataset_ext_swde}
\end{table}

\begin{table}
    \centering
    \small
    \begin{tabular}{lll}
    \toprule
        \textbf{Domain} & \textbf{Attribute} & \textbf{Website}\\
    \midrule
        \multirow{7}{*}{Book} & \multirow{7}{*}{\parbox{0.15\textwidth}{title \\ author\\ price}} & abebooks\\
        & & alibris\\
        & & barnesandnoble\\
        & & fishpond\\
        & & infibeam\\
        & & powells\\
        & & thriftbooks\\
        
        \midrule
        \multirow{7}{*}{E-commerce} & \multirow{7}{*}{\parbox{0.15\textwidth}{title \\ price}} & amazoncouk\\
        & & bestbuy\\
        & & dabs\\
        & & ebay\\
        & & pcworld\\
        & & tesco\\
        & & uttings\\

        \midrule
        \multirow{8}{*}{Hotel} & \multirow{8}{*}{\parbox{0.15\textwidth}{address \\ price \\ title }} & agoda\\
        & & expedia \\
        & & hotels\\
        & & hoteltravel \\
        & & javago\\
        & & kayak\\
        & & ratestogo \\
        & & venere \\
        
        \midrule
        \multirow{8}{*}{Movie} & \multirow{8}{*}{\parbox{0.15\textwidth}{actor \\ genre \\ title }} & 123movieto\\
        & & hollywoodreporter \\
        & & imdb\\
        & & mediastinger \\
        & & metacritic\\
        & & rottentomatoes\\
        & & themoviedb \\
        & & yidio \\
    \bottomrule
    \end{tabular}
    \caption{Detail statistic of \textsc{Ds1} dataset.}
    \label{tab:dataset_ds1}
\end{table}
\begin{table*}[h]
    \centering
    \small
    \begin{tabular}{lp{0.3\textwidth}p{0.55\textwidth}}
    \toprule
        \textbf{Domain} & \textbf{Task prompt} & \textbf{Prompt}\\
    \midrule
        \multirow{4}{*}{Auto} & \multirow{4}{*}{\parbox{0.3\textwidth}{Here's a webpage with detailed information about an auto.}} & Please extract the model of the auto.\\
        & & Please extract the price of the auto.\\
        & & Please extract the engine of the auto.\\
        & & Please extract the fuel efficiency of the auto.\\

        \midrule
        \multirow{5}{*}{Book} & \multirow{5}{*}{\parbox{0.3\textwidth}{Here's a webpage with detailed information about a book.}} & Please extract the title of the book.\\
        & & Please extract the author of the book.\\
        & & Please extract the isbn number of the book.\\
        & & Please extract the publisher of the book.\\
        & & Please extract the publication date of the book. \\

        \midrule
        \multirow{3}{*}{Camera} & \multirow{3}{*}{\parbox{0.3\textwidth}{Here's a webpage with detail information of camera.}} & Please extract the product name of the camera.\\
        & & Please extract the sale price of the camera.\\
        & & Please extract the manufacturer of the camera.\\

        \midrule
        \multirow{4}{*}{Job} & \multirow{4}{*}{\parbox{0.3\textwidth}{Here's a webpage with detailed information about a job.}} & Please extract the title of the job.\\
        & & Please extract the name of the company that offers the job.\\
        & & Please extract the working location of the job.\\
        & & Please extract the date that post the job. \\

        \midrule
        \multirow{4}{*}{Movie} & \multirow{4}{*}{\parbox{0.3\textwidth}{Here's a webpage with detailed information about a movie.}} & Please extract the title of the movie.\\
        & & Please extract the director of the movie. \\
        & & Please extract the genre of the movie.\\
        & & Please extract the MPAA rating of the movie.\\

        \midrule
        \multirow{4}{*}{NBAPlayer} & \multirow{4}{*}{\parbox{0.3\textwidth}{Here's a webpage with detailed information about an NBA player.}} & Please extract the name of the player.\\
        & & Please extract the team of the player he plays now.\\
        & & Please extract the height of the player.\\
        & & Please extract the weight of the player. \\

        \midrule
        \multirow{4}{*}{Restaurant} & \multirow{4}{*}{\parbox{0.3\textwidth}{Here's a webpage with detailed information about a restaurant.}} & Please extract the restaurant's name.\\
        & & Please extract the restaurant's address.\\
        & & Please extract the restaurant's phone number.\\
        & & Please extract the cuisine that the restaurant offers. \\

        \midrule
        \multirow{4}{*}{University} & \multirow{4}{*}{\parbox{0.3\textwidth}{Here's a webpage on detailed information about a university.}} & Please extract the name of the university.\\
        & & Please extract the contact phone number of the university.\\
        & & Please extract the website url of the university.\\
        & & Please extract the type of the university. \\
    \bottomrule
    \end{tabular}
    \caption{Prompts for scraper generation task in \textsc{Swde} dataset.}
    \label{tab:task_prompt}
\end{table*}

\lstset{
    basicstyle=\ttfamily,
    keywordstyle=\color{blue}\bfseries, 
    identifierstyle=\color{black},
    commentstyle=\color{gray},
    stringstyle=\color{red},
    escapeinside={(*@}{@*)},
    breaklines=true,
}

\begin{figure*}
\begin{tcolorbox}[title=Prompt of Top-down Operation, left=2mm,right=1mm,top=0.5mm, bottom=0mm,colback=white,colframe=blue!30!black]
\begin{lstlisting}[columns=flexible, numbers=none, xleftmargin=0mm, basicstyle=\small\ttfamily]
Here's the HTML extraction task:
Task description: Please read the following HTML code, and then return an Xpath that can recognize the element in the HTML matching the instruction below.
Instruction: {0} 
We will offer some history about the thought and the extraction result. Please reflect on the history trajectory and adjust the xpath rule for better and more exact extraction. Here are some hints:
1. Judge whether the results in the history are consistent with the expected value. Please pay attention to the following case:
    1) Whether the extraction result contains some irrelevant elements
    2) Whether the scraper returns an empty result
    3) The raw values containing redundant separators are considered consistent because we will postprocess them.
2. Re-thinking the expected value and how to find it depends on the xpath code
3. Generate a new or keep the origin xpath depending on the judgement and thinking following the hints:
    1. Do not output the xpath with the exact value or element that appears in the HTML.
    2. Do not output the xpath that indicates multiple nodes with different values. It would be appreciated to use more @class and [num] to identify the different nodes that may share the same xpath expression.
    3. If the HTML code doesn't contain suitable information to match the instruction, keep the xpath attribute blank.

Please output in the following JSON format:
{
    "thought": "", (*@\color{lightgray}{\# thought of why the xpaths in history do not work and how to adjust the xpath}@*)
    "consistent": "", (*@\color{lightgray}{\# whether the extracted result is consistent with the expected value, return yes/no directly}@*)
    "value": "", (*@\color{lightgray}{\# the value extracted from the HTML that matches the task description}@*)
    "xpath": "", (*@\color{lightgray}{\# a new XPath that is different from the XPath in the following history if not consistent}@*)
}

And here's the history of the thought, xpath and result extracted by scraper.
{1}

Here's the HTML code:
```
{2}
```
\end{lstlisting}
\end{tcolorbox}
\end{figure*}

\begin{figure*}
\begin{tcolorbox}[title=Prompt of Step-back Operation, left=2mm,right=1mm,top=0.5mm, bottom=0mm,colback=white,colframe=blue!30!black]
\begin{lstlisting}[columns=flexible, numbers=none, xleftmargin=0mm, basicstyle=\small\ttfamily]
Your main task is to judge whether the following HTML code contains all the expected values, which are recognized beforehand.
Instruction: {0}
And here's the value: {1}
The HTML code is as follows:
```
{2}
```

Please output your judgement in the following JSON format:
{
    "thought": "", (*@\color{lightgray}{\# a brief thinking about whether the HTML code contains expected value}@*)
    "judgement": "" (*@\color{lightgray}{\# whether the HTML code contains all extracted value. Return yes/no directly.}@*)
}
\end{lstlisting}
\end{tcolorbox}
\end{figure*}

\begin{figure*}
\begin{tcolorbox}[title=Prompt of Synthesis, left=2mm,right=1mm,top=0.5mm, bottom=0mm,colback=white,colframe=blue!30!black]
\begin{lstlisting}[columns=flexible, numbers=none, xleftmargin=0mm,basicstyle=\small\ttfamily]
You're a perfect discriminator who is good at HTML understanding as well. Following the instructions, there are some action sequences written from several HTML and the corresponding results extracted from several HTML. Please choose one that can be best potentially adapted to the same extraction task on other web pages on the same websites. Here are the instructions for the task:
Instructions: {0}
The action sequences and the corresponding extracted results with different sequences on different webpage are as follows:
{1}

Please output the best action sequence in the following JSON format:
{
    "thought": "" (*@\color{lightgray}{\# brief thinking about which to choose}@*)
    "number": " (*@\color{lightgray}{\# the best action sequence chosen from the candidates, starts from 0. If there is none, output 0.}@*)
}
\end{lstlisting}
\end{tcolorbox}
\end{figure*}

\end{document}